\documentclass[review]{elsarticle}

\usepackage {graphicx}
\usepackage{lineno,hyperref}
\usepackage{times}
\usepackage{epsfig}
\usepackage[table,xcdraw]{xcolor}
\usepackage{times}
\usepackage{amsmath}
\usepackage{amssymb}
\usepackage{multirow}
\usepackage{float}
\usepackage{lscape}
\usepackage[utf8]{inputenc}
\usepackage{footmisc}
\usepackage{pbox}
\usepackage{color, colortbl}
\usepackage{url}
\usepackage{csquotes} 
\usepackage{lscape} 
\newcolumntype{a}{>{\columncolor{gray}}c}

\journal{Computer Science Review}

\begin{document}

\hyphenation{FaceQnet}

\begin{frontmatter}

\title{Biometric Quality: Review and Application \\to Face Recognition with FaceQnet}


\author[mymainaddress]{Javier Hernandez-Ortega\corref{mycorrespondingauthor}}
\cortext[mycorrespondingauthor]{Corresponding author}
\ead{javier.hernandezo@uam.es}

\author[mysecondaryaddress]{Javier Galbally}

\author[mymainaddress]{Julian Fierrez}

\author[mysecondaryaddress]{Laurent Beslay}

\address[mymainaddress]{School of Engineering, Universidad Autonoma de Madrid, Spain}
\address[mysecondaryaddress]{European Commission, JRC, Italy}

\begin{abstract}
\textit{“The output of a computerised system can only be as accurate as the information entered into it.''} This rather trivial statement is the basis behind one of the driving concepts in biometric recognition: \textit{biometric quality}. Quality is nowadays widely regarded as the number one factor responsible for the good or bad performance of automated biometric systems. It refers to the ability of a biometric sample to be used for recognition purposes and produce consistent, accurate, and reliable results. Such a subjective term is objectively estimated by the so-called \textit{biometric quality metrics}. These algorithms play nowadays a pivotal role in the correct functioning of systems, providing feedback to the users and working as invaluable audit tools. In spite of their unanimously accepted relevance, some of the most used and deployed biometric characteristics are lacking behind in the development of these methods. This is the case of face recognition. After a gentle introduction to the general topic of biometric quality and a review of past efforts in face quality metrics, in the present work, we address the need for better face quality metrics by developing FaceQnet. FaceQnet is a novel open-source face quality assessment tool, inspired and powered by deep learning technology, which assigns a scalar quality measure to facial images, as prediction of their recognition accuracy. Two versions of FaceQnet have been thoroughly evaluated both in this work and also independently by NIST, showing the soundness of the approach and its competitiveness with respect to current state-of-the-art metrics. Even though our work is presented here particularly in the framework of face biometrics, the proposed methodology for building a fully automated quality metric can be very useful and easily adapted to other artificial intelligence tasks.
\end{abstract}


\end{frontmatter}


\section{Introduction}
\label{introduccion}

\blockquote{\textit{``On two occasions I have been asked, ‘Pray, Mr. Babbage, if you put into the machine wrong figures, will the right answers come out?’... I am not able rightly to apprehend the kind of confusion of ideas that could provoke such a question."} - Charles Babbage, Passages from the Life of a Philosopher, 1864.}

\medskip 
\textit{``Garbage In, Garbage Out."} The well-known computer science GIGO principle summarises in a very efficient and graphic manner, the pivotal role played by the soundness of the data given as input to any computerised system, in the meaningfulness of the output results. As in any other area of science, only nonsense conclusions can be expected from flawed premises. And automated digital systems are no exception to this rule.

In essence, the GIGO principle establishes a direct link between the reliability of the input and the output of a system. Therefore, it is easy to infer the huge advantages that would be brought about by a tool capable of assessing the robustness and accuracy of the input data to a specific automated system. Properly utilised, such a tool would have a major impact on the performance of the system, and on the ability of users to interpret its results, based on an objective measure of their consistency. In the field of biometrics, these invaluable tools are the so-called: \textit{biometric quality metrics}.

In biometrics, the general GIGO principle has been translated into the concept referred to as \textit{biometric quality}. Fundamentally, the simple underlying basis to biometric quality is that, if the biometric samples given as input to an automated recognition system are of low quality, unreliable inaccurate results will be generated. And, the other way around, if the acquired biometric samples are of high quality, low error rates will be achieved.

The previous \textit{biometric quality} statement leads to a foregone conclusion: high quality samples are preferable to low quality samples. However, such seemingly trivial assertion, raises one immediate fundamental question: How can biometric quality be measured so that we are capable of selecting high quality samples over low quality samples? That is, how can we define what biometric quality is? Furthermore, who establishes what a high/low biometric sample is? It is certainly not easy to give a closed, scientific, fact-based answer to these queries.

We all, as human beings, have an instinctive feeling of what a high or low quality sample is. For the sake of argument, let’s assume we are shown a well-focused frontal portrait of a person, with good homogeneous illumination, no shadows or occlusions, a uniform background, and high resolution. We would all agree that it represents a high-quality face image. Why? Because from such an image we would be able to recognise the person. On the other hand, a low-resolution facial picture, taken from an angle, somewhat blurred, and with heavy shadows, would be regarded by most as presenting low quality. Why? We would all have difficulties recognising the person based on that specific image. Analogue examples could be given for any other biometric characteristic.

In light of the previous argumentation, we may conclude that, in fact, we all possess a subjective, intuitive perception of biometric quality. However, it is difficult for us to translate this intangible insight into measurable objective numbers. While in the example presented above, we would all agree which  one is the high quality picture and which is the low quality one, each person would very likely assign to them a different quality measure from $0$ to $100$. This is, precisely, the huge challenge addressed by automated \textit{biometric quality metrics}: producing an objective quantitative estimation of an inherently subjective concept.

To this aim, quality metrics take advantage of a key notion hidden in the example proposed above regarding facial pictures. Ultimately, what makes humans decide whether or not a biometric sample is of high quality? Its ability to be used for its ultimate purpose: recognise the source of the sample. Unconsciously, the question being posed to ourselves to decide on the quality of a picture is: how likely is it that I would recognise the person based on this image? The same elementary principle can be exploited by automated quality metrics. They can assign a quality score to samples, based on the suitability of that sample for recognition purposes by automated systems. That is, a biometric quality metric can be, essentially, a predictor of biometric accuracy.

This understanding of biometric quality metrics as predictors of accuracy, is in line with the \textit{utility} definition given in standard ISO/IEC 29794 for the term quality \cite{REFstandard}. This is also, by far, the interpretation followed in most implementations described in the specialised literature, where quality metrics are being applied to a wide variety of tasks such as: quality control of large databases with multiple contributors, design of re-enrolment strategies in case of low quality acquisitions, quality-based multimodal fusion~\cite{phd_julian}, or adaptation of data processing techniques.

In spite of data quality being nowadays regarded as the number one factor impacting the performance of biometric systems \cite{grother2007performance,DraftREPORTnistFRVTQ}, the level of development and research effort in this field varies greatly among biometric characteristics. In particular, fingerprint recognition is clearly ahead in this unofficial classification, with countless published works and metrics~\cite{Alonso-Fernandez2007_FingerQualityReview}, that have led, perhaps most importantly, to the generation of a system-independent and open-source quality metric: the NIST Fingerprint Image Quality algorithm (NFIQ) \cite{tabassi2004fingerprint} (and NFIQ 2.0\footnote{https://github.com/usnistgov/NFIQ2}). This metric is widely accepted by the community as the golden standard that sets the performance bar for all other fingerprint quality algorithms. 

On the other hand, facial recognition is one of the most deployed biometric technologies with a great prospective market rise for the coming years~\cite{ranjan2018understanding}. However, with regard to the amount of effort devoted so far to data quality analysis, it is almost at the other end of the spectrum compared to fingerprints. At the moment, to meet the growth expectations raised by this technology, a point has been reached where it is necessary that face biometrics catches up with fingerprints in the study and understanding of data quality. This need has become ever more pressing by the advent of the new generation of biometric-enabled large European IT systems, such as the Schengen Information Systems (SIS), the Visa Information System (VIS) or the Entry Exit System (EES), which require of a normalised way to audit the quality of biometric data, shared by multiple contributors in central databases. In the current state of play, it is almost an unanimous claim coming from all stakeholders in the face recognition community (academia, governmental institutions, law enforcement agencies, border control agencies, and standardisation bodies), that a stronger commitment is required to work together on the generation of improved face quality metrics and, eventually, the development of a common standard benchmark similar to the fingerprint NFIQ.

While some valuable works have started to scratch the surface of face quality analysis (see Sect.~\ref{related_sect} for a review of the state of the art), there is still a long way to go before it reaches the level of progress exhibited by fingerprint-based systems, and before the requirements of the community are met. The current work is a solid step towards bridging this existing technological gap.

With this objective, the article presents an innovative approach to face quality assessment. The new method, FaceQnet, takes advantage of the largely demonstrated ability of deep learning networks to extract the most salient information from face images for recognition purposes. Through a knowledge transfer process, these machine-learned features are combined with training groundtruth quality scores, produced in a completely automatised way that does not involve human labelling. As a result, the system becomes fully scalable, not relying on the potentially biased human quality perception, but taking strictly into account for its training the one parameter which is expected to predict: the accuracy of automated recognition systems. Two successive versions of FaceQnet (v0 and v1) have been assessed and compared to other state-of-the-art methods, following our own evaluation and an independent evaluation performed by NIST, as part of their Face Recognition Vendor Test (FRVT) campaign. The results have shown an improvement between the two implementations of the algorithm, providing new insights into the problem of face quality, and proving the soundness and competitiveness of the approach.

\pagebreak

As ancillary contributions of the work, we provide FaceQnet as an open source project to the community\footnote{https://github.com/uam-biometrics/FaceQnet}, so that it can be used to further advance the field of face quality estimation. We have also generated and made available, together with FaceQnet, quality labels for popular face databases such as Labeled Faces in the Wild (LFW) and VGGFace2.


\begin{figure}[t!]
\begin{center}
\includegraphics[width=\linewidth]{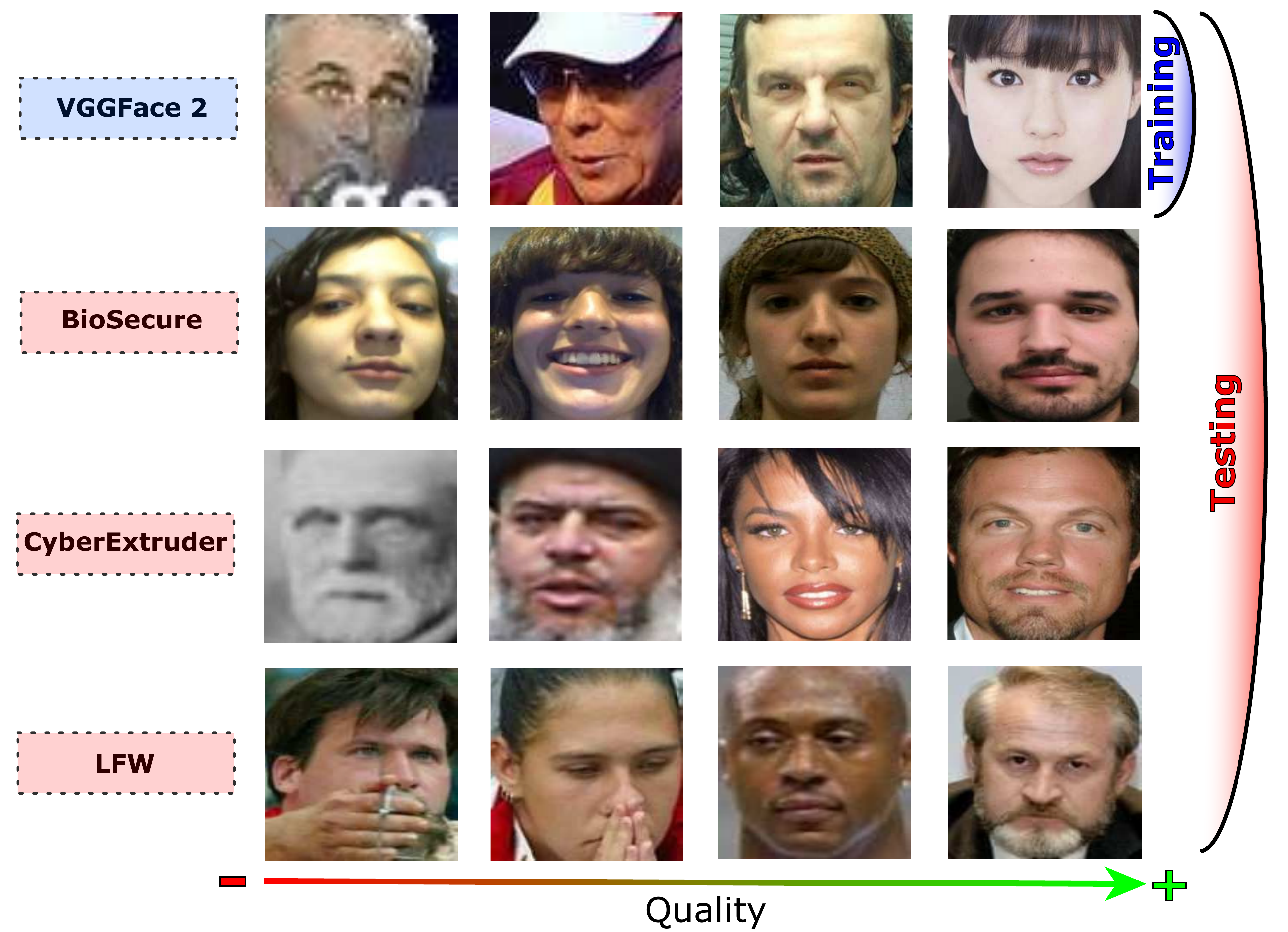}
\end{center}
\caption{\textbf{Examples of varying quality} from the four databases used in this paper: VGGFace2 \cite{cao2018vggface2}, BioSecure \cite{ortega2010multiscenario}, CyberExtruder, and LFW \cite{LFWTech}. The figure shows a selection of images from each database with variable quality according to their ICAO compliance values. The samples go from high quality images (right column) to low quality images (left column) which suffer from diverse variability factors such as low resolution, blur, bad pose, occlusions, etc.}
\label{VGGFace2}
\end{figure}

A preliminary version of FaceQnet was presented in \cite{2019_FaceQnet_Hernandez}, which will be referred to from now on as FaceQnet v0. Consequently, the model trained in the present work will be referred to as FaceQnet v1. 

The main contributions with respect to \cite{2019_FaceQnet_Hernandez} are: 1) a modification of the architecture of FaceQnet v0 to avoid overfitting; 2) the generation of new training groundtruth with data from more comparators, to reduce the system dependence of the quality measure; 3) an improved evaluation protocol, including a comparison with other metrics from the state of the art, and a larger variety of face images from four different public databases (see Fig.~\ref{VGGFace2}), in order to get a deeper knowledge of its accuracy regarding quality assessment for face recognition; and 4) a more comprehensive introduction and positioning with respect to related works. Even though FaceQnet is presented here in the framework of face biometrics, the proposed methodology for building an automated quality metric can be very useful for other problems in which a task performance prediction is desirable.

The rest of this paper is organised as follows. Section \ref{quality_measurement} provides an introduction to biometric quality, and its application in face recognition. Section \ref{related_sect} summarises related works in face quality assessment. Section \ref{framework_introduction} introduces the general framework for the development and evaluation of FaceQnet. Sections \ref{proposed} and \ref{evaluation} make a deeper description the development and evaluation stages, respectively. Finally, concluding remarks are drawn in Section \ref{conclusion_section}.

\section{Introduction to Biometric Quality Measures}
\label{quality_measurement}

In biometrics, a quality measure is essentially a function that takes a biometric sample as input and returns an estimation of its quality level \cite{2011_QualityBio_FAlonso}. That quality level is usually related to the \emph{utility} of the sample at hand, or in other words, the expected recognition accuracy when using that specific sample. Introducing high quality samples in a database should improve the accuracy of recognition systems, while low quality samples should have the opposite effect. 

The quality of samples can be also related to more subjective factors such as human perceived quality \cite{khodabakhsh2019subjective}. Other definitions of biometric quality are discussed in \cite{2011_QualityBio_FAlonso}: a quality measure can be an indicator of \emph{character}, i.e., properties of the biometric source before being acquired (e.g., distinctiveness); or can also be an indicator of \emph{fidelity}, i.e., the faithfulness of the acquired biometric sample with respect to the biometric source.

As in most of the related works in the literature, for the purpose of the present paper we concentrate on quality measures as predictors of recognition accuracy, i,e,. their \textit{utility}. This way, biometric quality measures must be related to recognition error rates like False Non-Match Rate (FNMR) and False Match Rate (FMR).


In face recognition we can distinguish between verification and identification. Face verification consists in a one-to-one comparison for validating the identity of a given probe image. In verification usually a reference image of high quality for each subject, captured in a controlled and collaborative scenario, is available. For this application of face recognition we can assume (if using an accurate face recogniser): 1) two high quality images of the same subject will produce a high verification score; 2) comparing a high quality image vs. a low quality image of the same person will produce a low verification score; and 3) false matches should produce low comparison scores independently of the quality of the samples. Summarizing, high quality images should obtain high match scores and low non-match scores, while low quality images should obtain low verification scores in both cases.

On the other hand, for identification, the probe face image is compared with all the identities stored in the reference database (one-to-many comparisons). In this case a high quality image is often not available, making necessary to know if a comparison in which both images are of low quality can output a high non-match identification score. Here, we theorise that the same quality factors that can cause false negatives (high FNMR) can also cause false positives (high FMR). Consequently, developing a quality measure based on FNMR should be enough for predicting recognition accuracy in both the verification and the identification cases.

Quality measures related to recognition accuracy can be categorised according to:

\begin{itemize}

\item \textbf{Groundtruth Definition}: One of the main differences between approaches for developing quality measures is the definition of high and low quality, i.e., the generation of the groundtruth. Some works employ \textit{human perception} as their groundtruth. Another approach consists in using an \textit{accuracy-based} groundtruth, which will result in a quality measure that represents the correlation between the input sample and the expected recognition accuracy of automatic systems.

\item \textbf{Type of Input}: Quality assessment modules can be also classified with respect to the amount of information they employ in order to obtain the quality measures. In a \textit{Full-Reference} approach (FR), a gallery sample with high quality is supposed to be available. The system compares the features from the probe samples with the ones from the high quality reference. In \textit{Reduced-Reference} methods (RR) just partial information of a high quality sample is available. \textit{No-Reference} methods (NR) do not use any reference information to compare with the probe sample. These methods apply prior information from the samples the system is dealing with, for example for building a statistical model.

\item \textbf{Features Extracted}: Biometric quality measures can also be classified in terms of the type of features that are extracted from the samples. Quality-related factors can be measured based on: 1) hand-crafted features, defined by the designer of the method based on past experience; or based on: 2) machine-learned features, e.g., generated by a Deep Neural Network (DNN) based on a pool of annotated training data.

\item \textbf{Output}: The output of the different quality assessment algorithms is not always the same, some methods may generate a qualitative label for each sample in the database in order to distribute the samples into a few quality ranges (e.g., low, medium, or high quality). Other methods just output a decision declaring if a specific sample is compliant with a quality standard or not. More complex works try to estimate the Probability Density Functions (PDFs) of the different variability factors present in the samples, e.g., blur or extreme pose. These PDFs will estimate the level of these quality factors in each sample of the database. Some of the most recent approaches compute a numerical score for each input sample (e.g., a real value in the range [$0$, $1$]), which serves as a predictor of the expected accuracy when using that sample for recognition.

\end{itemize}


The output of a biometric quality assessment module can be used at different stages of the recognition task. For example, it can be used during the enrollment process for giving feedback to the users or the operators. It can be also employed during the different recognition stages in order to improve the global accuracy of the comparator:

\begin{itemize}

\item \textbf{Photo acceptance/rejection}: Face quality can be used to reject an image if its quality is low, initiating the capture of a new image of higher quality. This is specially useful during the enrollment phase, when we want that to acquire samples of the highest possible quality.

\item \textbf{Selection of preprocessing techniques}: If a recognition system detects that a biometric sample does not present enough quality, it could activate some additional preprocessing techniques to improve the quality of the final sample. These techniques may involve significant computational overloads, so it is important to know when they can be useful \cite{new5,new6}.

\item \textbf{Context switching}: A single recognition framework may have different algorithms in its core, each one of them being robust against some specific variability factors and weak against others \cite{new7,new8}. The quality estimation module can help to switch between algorithms depending on the context at hand.

\item \textbf{Fusion at decision-level}: This case is closely related to context-switching. It consists in having several recognition algorithms, each one with its weaknesses and strengths. Instead of employing only one algorithm for recognition, they can all be employed in parallel, using the quality information to perform a smart fusion of their output scores, weighting each output according to the quality measure \cite{new9,new10}.

\item \textbf{Complementing features}: The quality measures can be considered as additional features for analysis and recognition algorithms. Incorporating them to the feature vectors can help to improve the accuracy of such analysis and recognition algorithms \cite{new11,new12}.

\item \textbf{Ensuring acquisition consistency}: Face images may be captured under very changing situations like a high number of subjects, various acquisition devices, and different illumination conditions. The output of this heterogeneous process is usually a collection of images with large quality variations. Quality measures can be used in this case to ensure the consistency of image quality, e.g., applying a threshold to grant a minimum quality or measuring the mean quality of the samples for revealing problems under specific acquisition conditions.

\item \textbf{Sample selection}: The quality information can be used for selecting only the best quality samples from a collection. Other approach consists in looking for samples in the database that have a similar level of variability than the probe sample. This way, the acquisition conditions from the gallery and the probe samples would be as close as possible, useful when the capture process includes any inevitable type of variability. This application of quality measures can boost the accuracy compared to using all the samples without taking their quality into account \cite{new13}.
 
\item \textbf{Template update/replacement}: When a subject is recognised with high enough confidence, the system could use the probe sample to improve or replace the template of that subject that is stored in the database \cite{new14,new15}.

\end{itemize}

\section{Face Quality Assessment: Related Works}
\label{related_sect}

Since the experimental part of the present article has been developed in the specific field of face recognition, in the present section we review the most relevant works that have been published so far proposing approaches for face quality estimation.

As already introduced from a general perspective in Sect. \ref{quality_measurement}, there are many application scenarios in which face recognition systems in particular, can take advantage of quality assessment. For example, in video-surveillance scenarios quality assessment can be employed for frame selection~\cite{2013_TIFS_PTome_SoftBiometrics}. In this type of settings, variability factors such as pose, occlusions, blur, etc, are usually present in the acquired images. As it has been stated previously, the recognition accuracy can be improved, for example, discarding the samples with low quality and using only the highest ones. In systems with strict storage requirements, the quality measures can be used to select the best quality images in order to store only those, reducing the amount required of storage. In forensic investigation, having a quality measure related to the face recognition accuracy may help to estimate the level of confidence of the decision~\cite{2015_FSI_FacialSoftBio_Pedro}. These are just a few examples of applications of face quality assessment.

To this date, a global standard on general face quality measure does not exist. However, several efforts have been done in that direction in the recent years. \cite{REFICAO} is an International Standard (IS) published by ICAO to define the characteristics of portraits on Machine Readable Travel Documents (MRTD). It is composed of a series of guidelines for the acquisition of high quality images (i.e., portrait-like), usually for their inclusion in official documents like ID cards or passports. This technical report is based on the ISO/IEC 19794-5:2005 and ISO/IEC 19794-5:2011 \cite{REFstandard2011}, and is now part of the ISO/IEC 39794-5 \cite{REFstandard2019}. Therefore, \cite{REFICAO} (and now the ISO/IEC 39794-5) are considered as the documents that define the ``perfect'' quality face picture, i.e., a ``portrait'', but they are not a general face quality standard. At the same time a number of vendors and academic works have developed their own tools to automatically check if an image complies with the guidelines given in these standards \cite{ferrara2012face}. In general, these works output a binary vector where each position defines whether or not a specific guideline was passed/not-passed by the image. 

A good example of the type of general quality standard we are talking about is the ISO standard on fingerprint image quality: ISO/IEC 29794-4 \cite{REFFinger}, that refers to the NFIQ developed by NIST as the standard fingerprint quality measure. Currently there are several works in progress for trying to reach this type of standardisation on face quality:

\begin{itemize}
\item The FRVT Quality Assessment campaign (FRVT-QA) held by US NIST\footnote{https://pages.nist.gov/frvt/html/frvt\_quality.html} is the first evaluation campaign aimed at comparing face quality metrics and set the current state of the art in the field.

\item Two new projects on face quality launched by ISO: the revision project ISO/IEC 297945-5, that wants to standarise the manner of evaluating the accuracy of face quality algorithms; and the ISO/IEC WD 24358 that wants to establish the requirements for automated systems that capture images of high quality faces.

\end{itemize}

\begin{landscape}

\begin{table*}[t!]
\caption{\textbf{Summary of quality assessment works} for face recognition, ordered chronologically and classified by: 1) the groundtruth definition process; 2) the type of input; 3) the features extracted; and 4) the type of output produced.}
\begin{center}
\resizebox{\linewidth}{!}{
\begin{tabular}{|c|c|c|c|c|c|}
\hline
\rowcolor{white}\textbf{Ref} & \textbf{Year} & \textbf{Groundtruth Definition} & \textbf{Type of Input} & \textbf{Features Extracted} & \textbf{Output}\\
\hline\hline

\hline

\cite{weber2006some} & 2006 & Human-based & No-Reference & Face features, image features & Score: individual presence of each factor\\
\hline

\cite{hsu2006quality} & 2006 & Human-based \& Accuracy-based & No-Reference & Face features, image features & Human perception score \& Machine recognition score \\
\hline

\cite{gao2007standardization} & 2007 & Human-based & No-Reference & Assymetric face features & Score: presence of each factor\\
\hline

\cite{abdel2007application} & 2007 & Human-based & No-Reference & Face features & Quality functions\\
\hline

\cite{beveridge2010quantifying} & 2010 & Human-based & No-Reference & Illumination & Individual score\\
\hline

\cite{abaza2012quality} & 2012 & Accuracy-based & Reduced-Reference & {Contrast, brightness, focus, sharpness and illumination} & FQI (Face Quality Index): $0$ to $1$\\
\hline

\cite{ferrara2012face} & 2012 & Human-based & No-Reference & 20 ICAO compliance features & Score from each individual test\\
\hline

\cite{phillips2013existence} & 2013 & Accuracy-based & Reduced-Reference & Image features, comparator features, sensor features
 & Low/high quality label\\
\hline

\cite{raghavendra2014automatic} & 2014 & Human-based & No-Reference & Texture features & Individual score\\
\hline

\cite{dutta2015predicting} & 2015 & Accuracy-based & No-Reference & 2 face features: pose, illumination  & Predicted FMR/FNMR\\
\hline


\cite{zhang2017illumination} & 2017 & Human-based & No-Reference & Illumination features & Individual score\\
\hline

\cite{wang2017learning} & 2017 & Human-based & No-Reference & $6$ image and face features & Individual score $1$ to $5$\\
\hline


\cite{best2018learning} & 2018 & Human-based \& Accuracy-based & No-Reference & CNN features & MQV (Machine-based Q.), and HQV (Human-based Q.)\\
\hline

\cite{khodabakhsh2019subjective} & 2019 & Human-based & No-Reference & Image features \& Ease of recognition & MOS: Mean Opinion Score \\
\hline

\textbf{FaceQnet v0 \cite{2019_FaceQnet_Hernandez}} & \textbf{2019} & \textbf{Human-based} \& \textbf{Accuracy-based} & \textbf{No-Reference} & \textbf{CNN features} & \textbf{Numerical quality measure: $0$ to $1$}\\
\hline

\cite{terhorst2020ser} & 2020 & Accuracy-based & No-Reference Unsupervised & CNN features & Numerical quality measure: $0$ to $1$\\
\hline

\textbf{FaceQnet v1 [Present paper]} & \textbf{2021} & \textbf{Human-based} \& \textbf{Accuracy-based} & \textbf{No-Reference} & \textbf{CNN features} & \textbf{Numerical quality measure: $0$ to $1$}\\
\hline

\end{tabular}
}
\end{center}
\label{related_works}
\end{table*}

\end{landscape}

One of the motivations behind the development of FaceQnet is to contribute to this standardisation effort, in order to reach a point similar to fingerprints, with NFIQ 2.0 as the standard go-to measure, defined in ISO/IEC 29794-4.

In Table \ref{related_works} we include a compilation of relevant related works in quality assessment for face recognition. The selection has been made to be representative of the different stages of face quality assessment research in the last $15$ years. In the following paragraphs we briefly describe relevant works in those research stages in chronological order.

First works related to face image quality assessment date back to early 2000's. The studies belonging to this first stage of research were generally centered in extracting hand-crafted features from face images and using them to calculate one or several quality measures. These features were meant to estimate the presence of one or various factors that have traditionally been considered to affect recognition performance, e.g., blurriness, non-frontal pose, or low resolution.

In \cite{weber2006some} the authors (workers from Cognitec, one of the most relevant companies in face recognition) presented one of the first compendia of quality measures and showed the relationship between those measures and the recognition performance of a Cognitec's face recogniser. The features they considered were all hand-crafted and included: the image sharpness, the openess of eyes, the pose, and the presence of glasses.

The work \cite{hsu2006quality} integrated several individual quality measures into an overall quality measure. That work computed various hand-crafted face-specific features like: lighting, pose, presence of eyeglasses, and resolution of the skin texture; and some image-specific features like: resolution of the complete image, existence of compression artifacts, and amount of noise coming from the acquisition sensor. The authors merged the individual quality measures into two different general measures: one based on human perception and other related to machine-recognition accuracy. They found that the quality measure related to machine-recognition was able to improve the recognition accuracy, while the correlation coefficient between the match scores and the human-based quality measure was much lower. According to the authors, that was because different persons gave different relevance to each individual quality measure, some of them not being critical for face recognition.

The research in \cite{gao2007standardization} presented a symmetry-based face quality assessment method that relied in the presence or absence of assymetries in the face. The authors considered that those assymetries can be caused by factors that have an impact in recognition performance, such as heterogeneous illumination and non-frontal pose.

The work presented in \cite{abdel2007application} introduced a quality assessment algorithm that checked the existence of factors like blur, heterogeneous lightning, non-frontal pose, and non-neutral expressions. The authors used eigenfaces for developing quality functions related to each of the different quality factors. However, they did not integrate the quality functions into a single measure for estimating the overall quality of a given face. 


One more of these ``classic'' hand-crafted approaches is the one presented in \cite{beveridge2010quantifying}, where the authors studied the effect that illumination has on face recognition, concluding that some of the best performing face recognition algorithms (at that time) were highly sensitive to different illumination levels when evaluating them with FRVT 2006.

In \cite{abaza2012quality} the authors proposed an accuracy-based Face Quality Index (FQI) combining individual quality factors extracted from five image features: contrast, brightness, focus, sharpness, and illumination. They used the CASPEAL database adding synthetic effects to the images (data augmentation), being able to emulate different real world variations. After computing a numerical quality value for each feature, they defined the Face Quality Index normalising each quality measure and modeling the distribution of quality measures as Gaussian PDFs. Values close to the mean of each PDF denoted high quality, while scores far to the mean represented low quality. The high quality reference PDFs were obtained using a high quality subset from the FOCS database. Finally they performed an average of all individual quality measures to compute the FQI. 

Another approach is described in \cite{ferrara2012face}. The authors presented the BioLab-ICAO framework, an evaluation tool for automatic ICAO compliance checking. The paper defined $30$ different individual tests for each input image. The output consists of a score for each test, going from $0$ to $100$. Even though this framework represents one of the first attempts of developing an automatic tool for knowing the level of compliance of an image with a public standard, it is not publicly available. Furthermore, the $30$ individual scores were not integrated into a final unified quality measure.

In \cite{phillips2013existence} the authors computed $12$ quality features divided into three categories. The first class consists of image processing and face recognition related features, e.g., edge density, eye distance, face saturation, pose, etc. The second category is composed of sensor-related features like the ones that can be encountered in the EXIF headers of the images. The last class consists of features related to the comparator they employed, i.e., SVM. They extracted conclusions about which features are more relevant to the specific dataset they used (PaSC) based on the overall recognition accuracy. They used that knowledge for splitting the whole dataset in two categories regarding quality: low and high.

The authors of \cite{raghavendra2014automatic} captured a database mimicking a real-life Automatic Border Control (ABC) scenario, and applied face quality assessment to its video sequences. ABC is probably one of the most relevant applications of face recognition, and improving its robustness is of great interest for the industry and for governmental institutions. That work evaluated the quality of the different frames of videos by analysing their texture and applied these quality measures for improving the recognition accuracy.

The work presented in \cite{dutta2015predicting} established a relationship between two image features, i.e., pose and illumination, and the final face recognition accuracy. They developed individual quality measures using PDFs in a way similar to \cite{abaza2012quality}. However, the main difference between both works is that in \cite{dutta2015predicting} the individual quality measures are employed to finally estimate expected accuracy values, i.e., False Match Rate (FMR) and False Non-Match Rate (FNMR). The authors used six different face recognition systems in order to extract accuracy values from the databases: two of them were Commercial Off-The-Shelf Software (COTS) and four were open-source algorithms, and they applied them to three different datasets: MultiPIE, FRGC and CASPEAL. Although their quality measure captured the relationship between pose and illumination, and face recognition accuracy, these are only two features among the high number of existing image quality variations.

The authors of \cite{zhang2017illumination} proposed a method based on a DCNN architecture for evaluating illumination quality values of given face images. They acquired their own Face Image Illumination Quality Database (FIIQD) and used it for training a ResNet-50 model (designed for object classification). Thus, the model is able of predicting quality illumination scores correlated with human perception.

In \cite{wang2017learning} the authors presented a learning-based facial image quality method that can be applied to select high quality frames inside a video sequence. They trained a random forest regressor to learn a subjective quality function using a subset of the LFW database labeled manually with quality scores going from
$1$ to $5$. 

With the recent growth of deep learning methods in face recognition due to their high accuracy, research works associated to face quality assessment are also adopting this type of approach successfully. For example, in \cite{best2018learning} the authors predicted quality measures related to recognition accuracy (referred to as Machine Quality Values, MQV) and other related to human perceived quality (Human Quality Values, HQV). They annotated the LFW database with human perceived quality using the Amazon Mechanical Turk platform where participants compared pairs of images from LFW and determined which one had the highest perceived quality. Differently to \cite{dutta2015predicting}, where they predicted a value for recognition accuracy, \cite{best2018learning} employed FMR and FNMR as accuracy values in the training stage and the output was a prediction of MQV or HQV. Other differential point of this work is that the authors employed a pretrained CNN (VGGFace) to extract features from the images. Then, they used those features to train their own classifier, which means that they successfully transferred knowledge from face recognition to quality prediction. The authors extracted interesting conclusions such as that both scores (MQV and HQV) are highly correlated with the recognition accuracy, even for cross-database predictions. They also concluded (based on their results) that automatic HQV is a more accurate predictor of accuracy than automatic MQV. The work in \cite{best2018learning} is probably one of the most advanced approaches to face quality estimation reported in the literature. However, it still presents some drawbacks: 1) a high amount of human effort is required to label the database with human perceived quality; and 2) a manual selection of a high quality image is needed for each subject to obtain the machine accuracy prediction, thus involving human effort and introducing human bias \cite{serna2020insidebias}. 


The work presented in \cite{khodabakhsh2019subjective} compared subjective and objective face quality measures and their effect on face recognition similarity scores. They asked $26$ participants to label face images with scores related with the ease of recognizing the face on the images. The authors then compared that subjective scores with other objective scores calculated using the guidelines of ISO/IEC TR 29794-5. They found that the correlation of the subjective scores to the recognition scores outperformed the objective scores.

In \cite{2019_FaceQnet_Hernandez} we presented FaceQnet v0, a deep learning method that had the objective of correlating the quality of an image to its expected accuracy for face recognition. It was designed as an extension of the work presented in \cite{best2018learning}. We employed the BioLab-ICAO framework \cite{ferrara2012face} for labeling the images of the VGGFace2 database with quality information related to their ICAO compliance level. The training of FaceQnet v0 was done using that automatically labelled groundtruth. We showed that the predictions from FaceQnet v0 were highly correlated with the face recognition accuracy of a state-of-the-art commercial system. However, our proposal had some limitations: we used only one face recogniser for the groundtruth generation (probably introducing system dependence); the presence of outliers in the groundtruth data affected significantly the training process; and because our testing protocol only included two different databases, we were not able to extract conclusions that could be applied to other data with entire confidence.

Some recent face quality assessment works already mention FaceQnet v0 among their main references. One of them is \cite{terhorst2020ser}, in which the authors proposed a face quality assessment method based on unsupervised learning. They computed the variations in the face embeddings coming out from several CNNs pretrained for face recognition. They developed a quality indicator by measuring the robustness across the different embeddings for a single face image. The authors compared their solution against six state-of-the-art face quality assessment approaches (being FaceQnet v0 among them).


%

Works in face quality assessment have followed a parallel path to those of face recognition~\cite{ranjan2018understanding}. In both cases, first works were based on hand-crafted features, designed by researchers following their intuition about which factors may be more relevant to the problem to solve. Also similarly to face recognition, the most promising current approaches are those based on deep learning. This type of algorithms have shown to outperform hand-crafted approaches as long as there are enough training data available. Using these data, deep learning models are able to infer the relationships between inputs and expected outputs, even when they are non-linear. However, the lack of labelled data, e.g., in face quality assessment, makes really difficult to train deep models accurately. Summarizing, thanks to this review of face quality assessment works we have identified the pressing need of a method for facilitating the task of labelling training data with quality values.

We have also identified the worrying lack of a common definition of face quality. Each related work defines its own vision of what a quality measure should be, from the selection of what variability factors must be taken into account, to the class of the final quality measure (sometimes a number, others a label, etc). We think that face quality assessment should follow a similar way to what has been done with fingerprint quality, i.e., the NFIQ 2.0 algorithm developed by NIST, that has been adopted as a golden standard.

The present work represents a step forward in overcoming the limitations of \cite{best2018learning} and FaceQnet v0 \cite{2019_FaceQnet_Hernandez}. As a result, our proposed solution, i.e., FaceQnet v1 is: 1) based on state-of-the-art deep learning; 2) massively scalable without human intervention, thanks to the fully automatic generation of the groundtruth quality labels; 3) developed and tested using multiple face datasets and state-of-the-art face recognition systems; and 4) validated in an independent evaluation by NIST.

\section{FaceQnet: Overview}
\label{framework_introduction}

\begin{figure}[t!]
\begin{center}
\includegraphics[width=\linewidth]{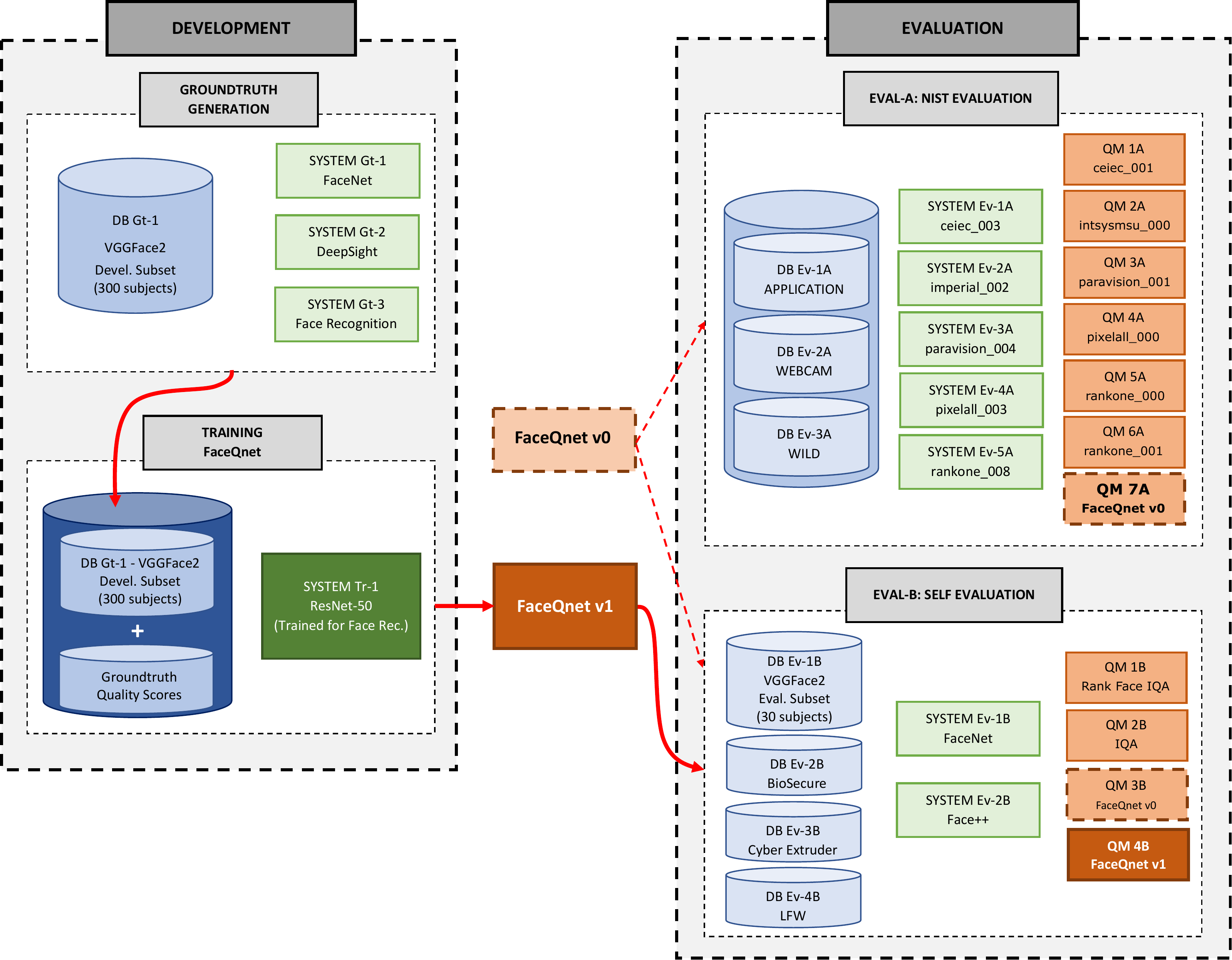}
\end{center}
\caption{\textbf{General scheme of FaceQnet's experimental framework}. The left part of the figure shows the development stage of FaceQnet v1: i) the generation of groundtruth quality measures, and ii) the training of the deep regression model. The right half of the figure describes the two evaluations included in this paper: i) an external assessment made by NIST (FVRT-QA) in which we participated with a preliminary version of FaceQnet (v0), and ii) a self-evaluation of FaceQnet v1 following the same protocol than in NIST evaluation. The figure indicates the databases (DB), face recognisers (SYSTEM), and other QA methods (QM) that we used at each point of the development and evaluation stages.}
\label{big_figure}
\end{figure}

The current work is divided in two different stages: a development phase in which we build and train FaceQnet v1 and a evaluation phase where we apply FaceQnet v1 and other quality measures from the literature to different databases comparing their accuracy in face quality assessment. Fig.~\ref{big_figure} shows the general scheme of FaceQnet's development and evaluation framework, and the different resources in the form of databases, face recognisers and quality measures that have been used in each one of the stages.


The left half of Fig.~\ref{big_figure} depicts the development stage in which we first generated a quality groundtruth by labeling a training database ($300$ users of VGGFace2) with automatically generated quality measures based on ICAO compliance using recognition scores from three different face recognisers. The second part of the development section consisted in training the new FaceQnet model, FaceQnet v1, with the quality groundtruth generated in the previous step. A deeper explanation is given in Sect.~\ref{proposed}, describing the complete process, the databases, and the systems.

The right half of Fig.~\ref{big_figure} describes the evaluation stage. In this work we first include the results of an external evaluation made by NIST for measuring the accuracy of face quality assessment methods, i.e., the Face Recognition Vendor Test - Quality Assessment (FVRT-QA)~\cite{DraftREPORTnistFRVTQ}. We took part in the evaluation with a preliminary version of FaceQnet (v0). Participating in that competitive test has been useful to show the soundness of our approach and to identify its potential strengths and weaknesses.

The second evaluation included in this paper consisted in a self-assessment of FaceQnet v1, in which we followed a similar protocol than in NIST evaluation, but using other evaluation databases and face recognisers. In this assessment we also compared FaceQnet v1 to other quality measures, some designed for face quality assessment and others for general image quality assessment. The protocols, databases, and systems used in both evaluations are described extensively in Sect. ~\ref{evaluation}.

\section{FaceQnet: Development}
\label{proposed}

Biometric quality estimation can be seen as a prediction of biometric accuracy, i.e., a regression problem. With FaceQnet we solve this regression problem in a supervised way using a groundtruth database composed of pairs of face images and their corresponding groundtruth quality measures.

The present paper is based on our preliminary work in \cite{2019_FaceQnet_Hernandez}, FaceQnet v0, whose objective was to correlate the quality of an image to its expected accuracy for face recognition. In the present work, we extend the results obtained in \cite{2019_FaceQnet_Hernandez} by improving its main weak points: 
1) the quality groundtruth is now generated using three different face comparators instead of only one; 2) the learning architecture is revised in order to avoid overfitting; and 3) the testing protocol now includes an external evaluation of FaceQnet v0 performed by NIST as part of FRVT-QA and a comparative evaluation over four different databases of varying quality (instead of only two), making both evaluations possible a deeper understanding of how FaceQnet v0 and v1 work. 

\subsection{Generation of the Groundtruth Quality Measures}
\label{groundtruth_gen}


We can think of quality in face recognition as a measure of the intra-class variability of a subject's images. The ICAO technical report \cite{REFICAO} imposes very strict guidelines when capturing new images. Controlling variability factors such as resolution, illumination, pose, focus, etc~\cite{2018_TIFS_SoftWildAnno_Sosa}, makes the images coming from the same subject to look as similar as possible, i.e., low intra-variability. This way, the comparison scores should be only dependent of the differences between different subjects, i.e., inter-class variability. Based on that rationale, in the current work we have made the next hypothesis in order to compute the quality groundtruth:

\begin{itemize}
    \item \textbf{HYPOTHESIS 1}: In this work we make the assumption that a perfectly compliant ICAO image represents perfect quality due to its low intra-class variability. Therefore, we assume that the mated comparison score between such perfect quality picture $A$ (i.e., ICAO compliant) and a picture $B$ of unknown quality can be a valid and accurate reflection of the quality level of picture $B$ (its level of intra-class variability). If the comparison score is low, this must be due to the low quality of the image $B$, since $A$ is of known perfect quality. On the other hand, if the score is high, it can be assumed that the second image is of good quality, containing a low level of variability factors such as the ones mentioned before. Therefore, that comparison score can be used as a machine-generated groundtruth quality for picture $B$.
\end{itemize}

This hypothesis has been made previously in other works related with biometric quality such as \cite{grother2007performance} and \cite{DraftREPORTnistFRVTQ}. These papers wanted to do quality assessment based on recognition accuracy using only a single biometric sample. Since recognition scores depend of both samples being compared, this assumption shown to be a robust alternative for developing no-reference quality assessment methods.

To know which images from the training database were closest to ICAO compliance, we used the BioLab framework from \cite{ferrara2012face}. This framework outputs a score between $0$ and $100$ for each one of its $30$ individual ICAO compliance tests. Not all of these tests have the same relevance for face recognition, so we selected a subset of them and then we computed a final averaged global ICAO compliance value. More specifically, the tests that we have selected are: blur level, too dark/light illumination, pixelation, heterogeneous background, roll/pitch/yaw levels, hat/cap presence, use of glasses, and presence of shadows.

As the training set for our quality assessment measure, we selected a subset of $300$ subjects from the VGGFace2 database \cite{cao2018vggface2} which will be referred to from now on as DB Gt-1. This database contains $3.31$ million images of \mbox{$9$,$131$} different individuals, with an average of $362.6$ images for each subject. All the images in the database were obtained from Google Images and they correspond to well known celebrities such as actors/actresses, politicians, etc. The images were acquired under unconstrained conditions and present large variations in pose, age, illumination, etc. These variations imply different quality levels. In the evaluation section of this work we also use another disjoint subset from VGGFace 2 composed of $30$ subjects to check the accuracy of our quality measure. Examples of images with different qualities from VGGFace 2 can be seen in Fig.~\ref{VGGFace2}.

For each subject in the training set we selected the image with the highest ICAO compliance value as the reference image, and we used the rest as probe images. To obtain the comparison scores that, according to the rationale explained in hypothesis 1, will be used as our groundtruth quality scores, we input each probe image of the training database into three face recognisers to extract three different $128$-dimensional feature vectors. We decided to use a varied selection of open-source and proprietary state-of-the-art recognisers in order to develop a quality groundtruth as less system-dependent as possible: 

\subsubsection{FaceNet \cite{schroff2015facenet} (SYSTEM Gt-1):} An open-source deep-learning model published by Google. Since its publication FaceNet has been used extensively in face recognition research, obtaining results of proven accuracy. It also has the advantage of having implementations in the most popular programming frameworks like Tensorflow or Pytorch. 

\subsubsection{DeepSight \cite{DeepSight} (SYSTEM Gt-2):} A proprietary software developed by BaseApp, complements our selection of comparators from a commercial point of view. It consists of an API interface that makes requests to a \textit{http} service, so it can be run both locally or in the cloud. It contains deep learning models for face detection, gender/age classification, and face recognition (we have only used the last one).

\subsubsection{Face Recognition \cite{face_recog} (SYSTEM Gt-3):} An open-source state-of-the-art face recogniser built using the \textit{dlib} library. We decided to use this recogniser because of its very good balance between performance (it shown an accuracy of 99.38\% on the Labeled Faces in the Wild benchmark), and ease of use thanks to its implementation of a simple command line tool for making face comparisons.

Using the embeddings provided by the face recognisers, we computed the Euclidean Distance between each reference image and all the remaining samples of the same subject. These distances represent the dissimilarity between each test image and its correspondent ``ICAO compliant" template. This process gave us three different mated distances for each pair of images. In order to fuse the distances into only one (used as the training groundtruth), they have been transformed to similarity scores $s$ into the [$0$,$1$] range using the next equation: $s = 1/(1 + e^{d})$, being $d$ each mated distance with zero mean and unitary standard deviation. 



\begin{figure}[t!]
\begin{center}
\includegraphics[width=\linewidth]{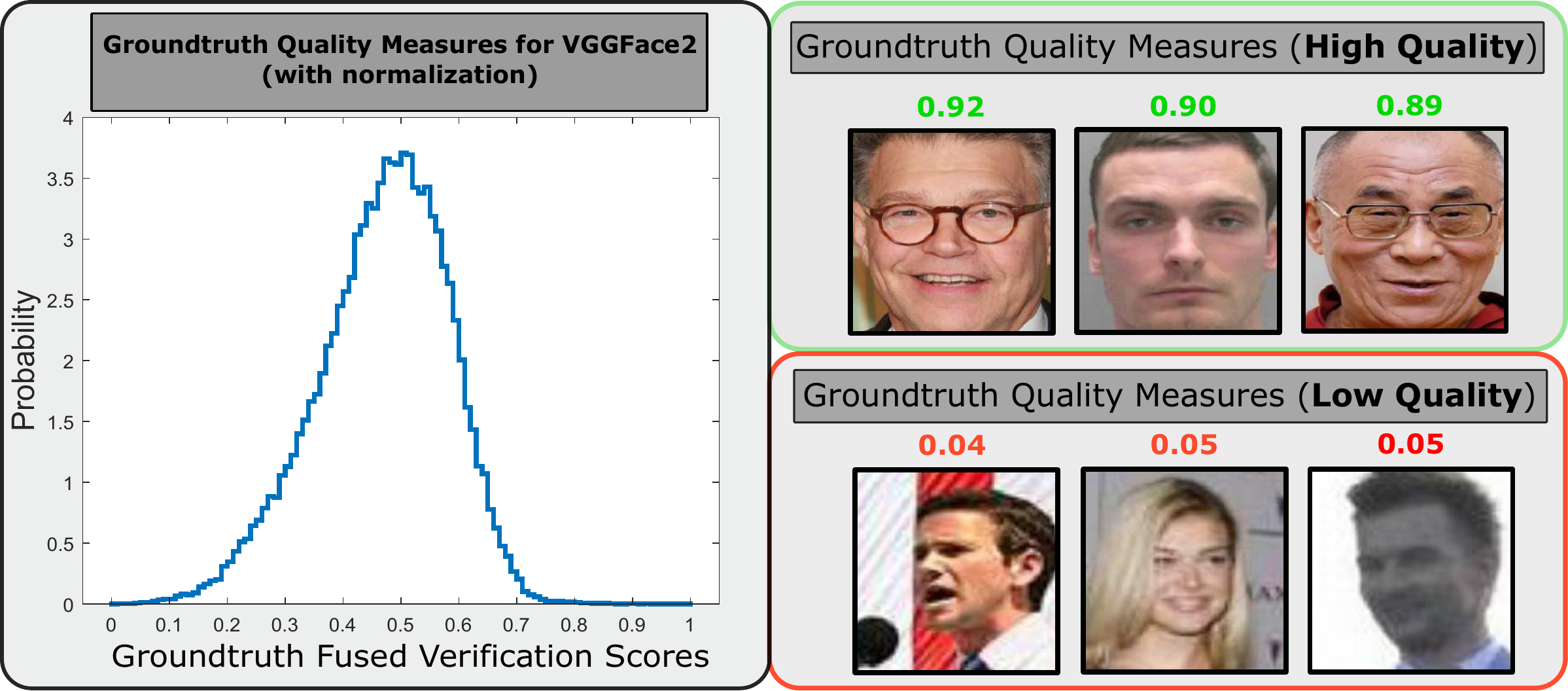}
\end{center}
\caption{\textbf{(Left) Distribution of the groundtruth quality measures for training FaceQnet v1.} The training quality measures are a combination of the verification scores obtained using FaceNet, DeepSight, and Face Recognition. \textbf{(Right) Training images of high and low subjetive quality}, and their corresponding groundtruth quality scores.}
\label{groundtruth_scores}
\end{figure}

Finally, the three normalised similarity scores were averaged to obtain the final groundtruth quality measures for training FaceQnet v1. As explained above in hypothesis 1, given that the reference images used to compute the similarity scores are nearly ICAO-compliant images of ``perfect'' quality we can assume that the final similarity score represents the quality level of the probe image. If the resultant similarity scores are high, the correspondent probe images are likely to have high quality characteristics. On the contrary, if the scores are low we can assume that the probe images have low quality regarding the face recognition task. 

One of the most relevant improvements here respect to FaceQnet v0 is using three different face recognisers instead of only one, for trying to avoid the quality groundtruth (the verification scores) to be system dependent. If we employed just one single recogniser, the resulting quality measure would be highly accurate when estimating the recognition performance of that training matcher, but it might not be useful for predicting the accuracy of recognisers never seen before.

Fig.~\ref{groundtruth_scores} (left) shows the distribution of the fused verification scores we used as the groundtruth quality measures for training FaceQnet v1. We calculated verification scores using the FaceNet, Face Recognition, and DeepSight recognisers, we normalised them to the [$0$,$1$] range, and then we combined them into a final groundtruth quality measure. Fig.~\ref{groundtruth_scores} (right) shows some examples of training images of high and low subjective quality (selected manually) and their associated groundtruth quality measures. The figure shows that the measures are correlated to the subjective quality of the images, i.e., its level of ICAO compliance. With the experiments included in Sect. \ref{evaluation} we prove that the quality measures are also related to face recognition accuracy.

\subsection{Training of the Deep Regression Model}

The proposed model, FaceQnet v1, is able to return a reliable prediction of the face recognition accuracy using just a probe image as its input. To that end, it performs end-to-end regression for quality estimation. Due to the limited amount of face quality training data, we opted to apply knowledge-transfer~\cite{patel2015visual}, which has been shown to be very effective in other face analysis problems such as gender~\cite{2021_TPAMI_SensitiveNets_Morales} or emotion estimation~\cite{2020_ICPR_Emotional_Pena}. In these works, the authors used a model pretrained for a different (but closely related) task using massive data, and they retrained it using only a limited set of data of the target task~\cite{Fierrez-Aguilar2005_AdaptedMultimodal}. This observation led us to the next hypothesis:

 \begin{itemize}
    \item \textbf{HYPOTHESIS 2}: Facial feature vectors containing information about \textit{identity} are quite likely to also contain information regarding face \textit{quality}, given the close relationship that exists by definition between recognition accuracy (i.e., identity) and quality (in terms of character of the source, fidelity of the capture device and utility of the sample). Therefore, using knowledge transfer we should be able to extract quality-related information from feature vectors that were initially designed for recognition purposes.
\end{itemize}

 To use face-recognition embeddings for quality estimation, we need to extract the quality related information from them, and this is done by using the groundtruth quality measures described in Sect.~\ref{groundtruth_gen}. The creators of the VGGFace2 database also published a CNN based on the ResNet-50 architecture \cite{he2016deep} pretrained with their database, showing that they were able to obtain state-of-the-art results when testing against challenging face recognition benchmarks such as IJB-C \cite{maze2018iarpa}, QUIS-CAMPI \cite{neves2015quis,2017_IntelligentSystems_icb-rw}, or PaSC \cite{beveridge2013challenge}. This is the model we used as the basis of FaceQnet v0 and v1, applying knowledge-transfer to change its domain from face recognition to quality assessment. This model will be referred to from now on as SYSTEM Tr-1. We removed the last classification layer of the base model and we substituted it with two additional Fully Connected (FC) layers to perform quality estimation. SYSTEM Tr-1 extracts a vector of \mbox{$2$,$048$} elements designed for face recognition. The first added FC layer combines the elements of the embeddings, synthesising them into feature vectors of $32$ elements that concentrate the quality related information. The second FC layer performs a final regression step that outputs a score, i.e., the final quality measure that helps us to know the level of suitability of an image for face recognition.
 
\begin{figure}[t!]
\begin{center}
\includegraphics[width=\linewidth]{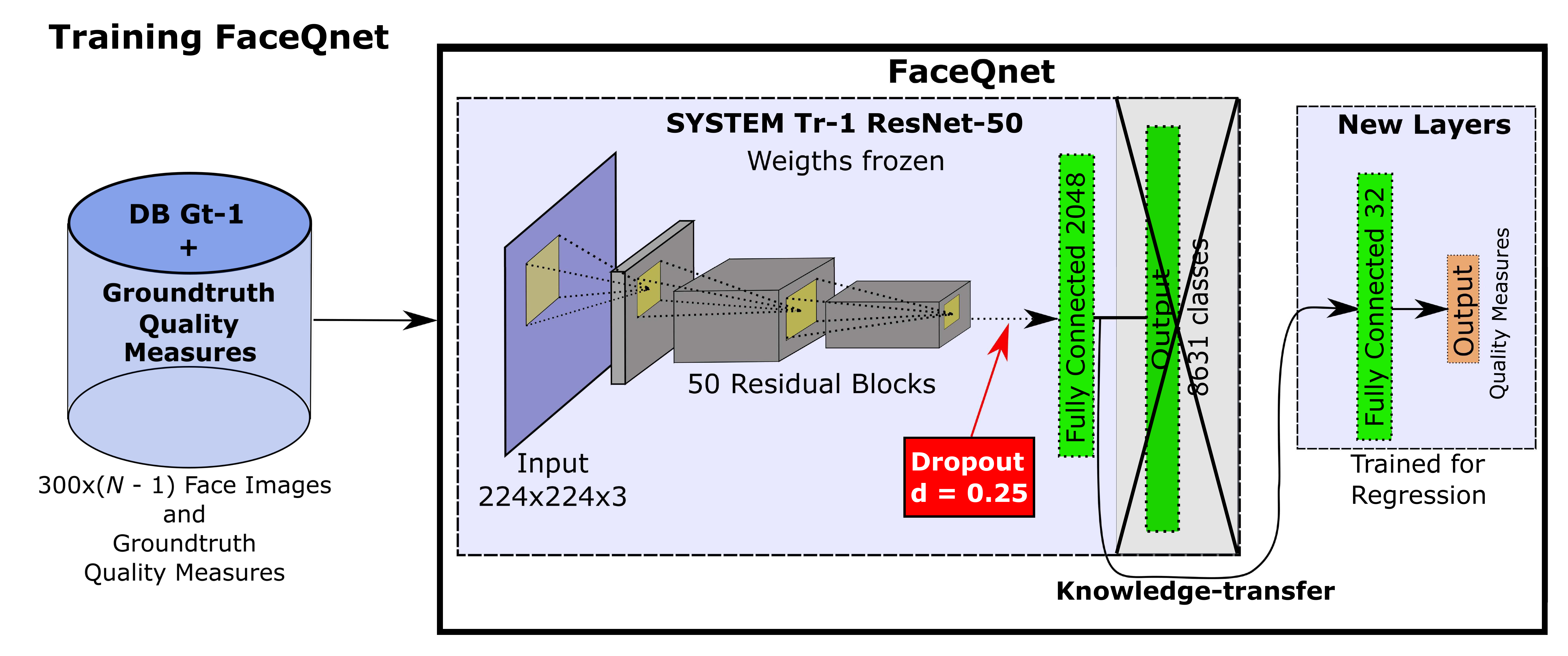}
\end{center}
\caption{\textbf{FaceQnet} (both v0 and v1 versions) is originally based on the ResNet-50 architecture \cite{he2016deep}, but replacing the last classification layer with two new ones designed for quality regression. For FaceQnet v1 we also added a dropout layer before the first additional fully connected layer. We trained only the new layers keeping the weights of the rest frozen, using the training set of face images and their groundtruth quality measures.}
\label{general_scheme_2}
\end{figure}


In order to improve the preliminary results from FaceQnet v0 \cite{2019_FaceQnet_Hernandez}, in FaceQnet v1 we also added a dropout layer before the first FC layer to avoid overfitting and generalise better when facing images from different datasets and scenarios. The final architecture of FaceQnet v1 is shown in Fig.~\ref{general_scheme_2}. In addition to the changes made into the generation of the groundtruth data, where we used three different comparators for avoiding system dependence, this change in the architecture makes the model more system and data independent in comparison to FaceQnet v0~\cite{2019_FaceQnet_Hernandez}.

The input to the network are face images of size $224\times224\times3$ previously cropped and aligned using MTCNN \cite{zhang2016joint}. We froze all the weights of the old layers and we only trained the new layers using the quality groundtruth generated in the previous step (see Sect. \ref{groundtruth_gen}). 

Once trained, FaceQnet can be used as a ``black box" that receives a face image and outputs a quality measure between $0$ and $1$ related to the face recognition accuracy. This quality measure can be understood as a proximity measure between the input image and a hypothetical corresponding ICAO compliant face image. The general biometric quality standard ISO/IEC 29794-1:2016 determines/stipulates that quality measures should produce continuous quality scores in the range [$0$-$100$] in contrast to discrete quality values or quality classes. FaceQnet complies with this standard since continuous values in the range [$0$-$1$] are directly convertible to the range [$0$-$100$].

\section{FaceQnet: Evaluation}
\label{evaluation}

\subsection{US-NIST Assessment: FaceQnet v0}
\label{nistevaluation}

As part of their Face Recognition Vendor Test, the US NIST started in 2019 an on-going evaluation of face quality metrics, the FRVT Quality Assessment\footnote{https://pages.nist.gov/frvt/html/frvt\_quality.html}. To date, there has been one wave of algorithms assessed in the competition. This first campaign comprised eight algorithms coming from six different participants and included the initial version of FaceQnet (v0) described in the preliminary work \cite{2019_FaceQnet_Hernandez}. A description of the objectives, experimental protocol, and the full results of the competition so far, were recently presented in a technical report \cite{DraftREPORTnistFRVTQ}.

\begin{figure}[t!]
\begin{center}
\includegraphics[width=\linewidth]{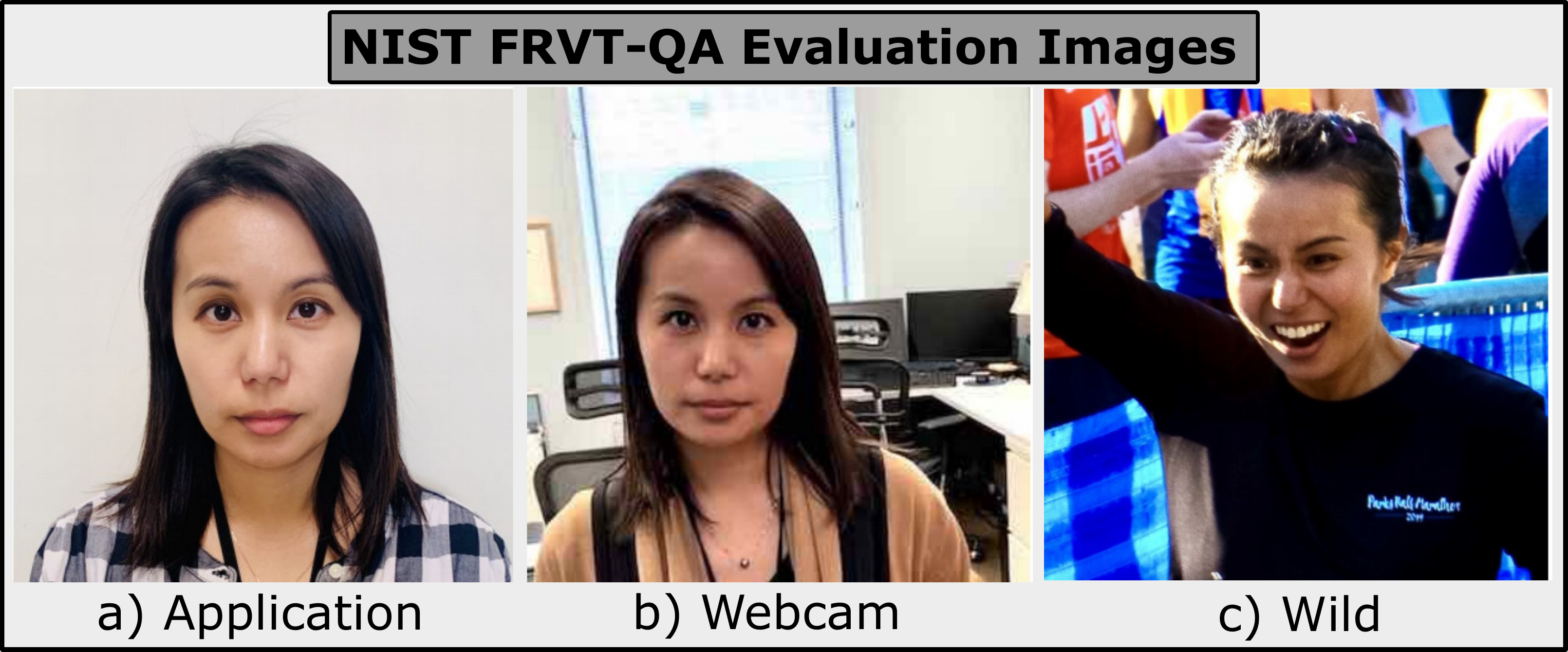}
\end{center}
\caption{\textbf{Samples of images used in the NIST assessment~\cite{DraftREPORTnistFRVTQ}.}}
\label{nist_images}
\end{figure}

The FRVT-QA evaluation was performed over a database that contained, for all subjects, three different image categories, each of them with a different expected quality level: 1) ``Application" pictures, which correspond to high-resolution ICAO-type portraits (very high quality); 2) ``Webcam" pictures, which correspond to close-to frontal images, taken indoors with a cooperative subject and no specific control over illumination or distance to the camera (good-to-average quality); 3) ``Wild" pictures, which include photojournalism-style photos, taken under unconstrained conditions with large variations in resolution (large quality range, from very poor to good). Samples of the types of images used in the NIST evaluation are shown in Fig.~\ref{nist_images}.



The evaluation included two main types of results for all the algorithms assessed: 1) Quality score distributions for each of the three image categories (i.e., application, webcam and wild); and 2) Error versus Reject Curves (ERC) for different comparators for two verification tasks: ``Application vs Webcam" and ``Wild vs Wild". 

ERC plots (defined in detail in \cite{grother2007performance}) are the most widely and commonly accepted metric to evaluate the performance of biometric quality measures. Note that ERC curves are computed taking only into account mated scores given that, as was already explained in the introduction of Sect.~\ref{quality_measurement}, quality measures are, ultimately, estimators of the variability among samples of the same user. ERC plots are calculated by discarding an increasing amount of images with low quality measures and then obtaining the new values of the FNMR. The initial threshold is set to fix the desired initial value of the FNMR by using the quantile function with the mated compared pairs. The same threshold is used for all the values in each ERC plot. 

\begin{figure}[t!]
\begin{center}
\includegraphics[width=0.9\columnwidth]{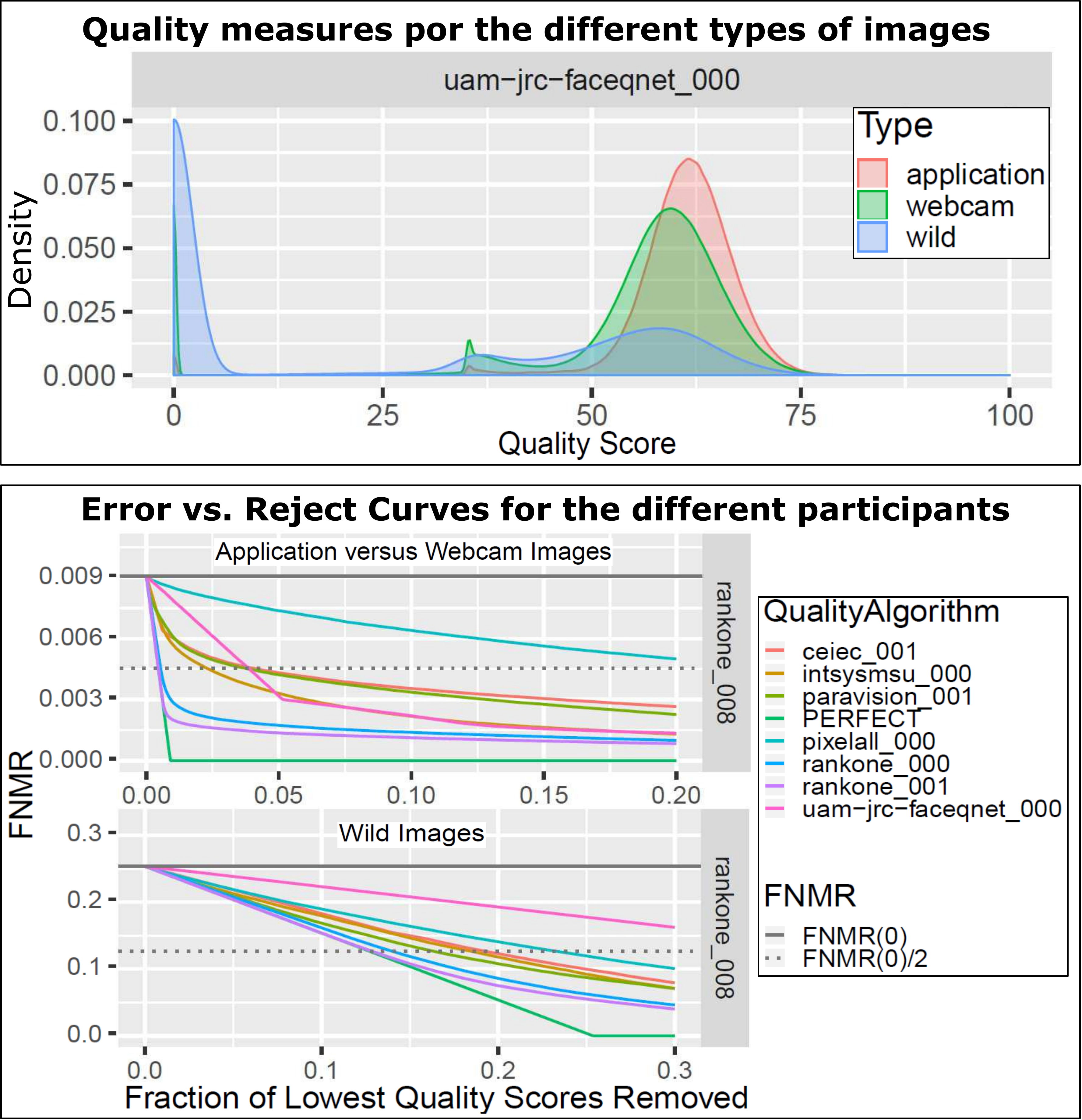}
\end{center}
\caption{\textbf{Brief summary of the results from the FRVT-QA campaign organised by NIST for the evaluation of face quality metrics.} The graphs have been directly extracted from \cite{DraftREPORTnistFRVTQ} and show the performance of the first version of FaceQnet presented in \cite{2019_FaceQnet_Hernandez} (FaceQnet v0). \textbf{(Top) quality measures} for the three different types of images in the evaluation database. \textbf{(Bottom) ERC plots} showing the performance of the different quality assessment methods submitted to the evaluation. For a full description of the competition and results, we refer the interested reader to \cite{DraftREPORTnistFRVTQ}. (Color image)}
\label{nist_eval}
\end{figure}

The curves show the relationship between the FNMR and the reject rates, describing how the FNMR (ideally) decreases when the data with the worst quality is discarded. An ideal quality measure would present a perfect correlation with recognition accuracy, which would entail that its ERC curve would coincide with the FNMR of a given recognition system. In the ERC plots shown in Fig. \ref{nist_eval} (bottom), Fig. \ref{ERC_facenet}, and Fig. \ref{ERC_faceplusplus} the curves designed as PERFECT represent that perfect correlation. Therefore, for different quality measures, the closer the ERC to the PERFECT curve, i.e., the ideal behaviour, the more accurate is the quality measure.


A brief summary of the FRVT-QA results is shown in Fig.~\ref{nist_eval}. In \cite{DraftREPORTnistFRVTQ} they used several different recognisers that participated in their Face Recognition Vendor Test. In this paper we include the results obtained using the rankone-008 recogniser of Rank One Computing\footnote{https://www.rankone.io} (SYSTEM Ev-5A rankone\_008 in Fig. \ref{big_figure}) as a summary of the whole evaluation. However, the complete results of the assessment can be checked in the full report. The graphs in the figure have been directly extracted from \cite{DraftREPORTnistFRVTQ} and have been selected to reflect the performance of the initial version of FaceQnet submitted to the evaluation \cite{2019_FaceQnet_Hernandez} (FaceQnet v0). Fig.~\ref{nist_eval} (top) depicts the quality score distributions of FaceQnet v0 for the three image categories. The graph in Fig.~\ref{nist_eval} (bottom) show the ERC plots for all the algorithms in the evaluation, based on the mated comparison scores obtained with the comparator ``rankone\_008", for a FMR of $0.1\%$, both for the ``Application vs Webcam" scenario and the ``Wild vs Wild" scenario.  

The main conclusions that can be extracted from these results are:

\begin{itemize}
    \item Given the quality score distributions shown in Fig.~\ref{nist_eval} (top), we can say that FaceQnet v0 is capable of distinguishing with a reasonable accuracy the difference in quality present in the three image categories considered in the competition. However, it has a tendency to saturate on the low-end of the quality range, that is, it has a significantly limited ability to discern between poor quality images, assigning to all of them very low quality values (see the abnormal high lobe of the wild distribution around quality value $0$).
    
    \item Fig.~\ref{nist_eval} (bottom) shows that FaceQnet v0 performs reasonably well in the quality estimation of average, good, and very high quality images (i.e., webcam and application categories). This could already be noticed in the distributions shown in the top graph and is further confirmed by the ERC curves of the ``Application vs Webcam" scenario, where, for most of the curve, FaceQnet v0 only performs worse than the two ``rankone" quality metrics. Please note that these ERC curves have been extracted using a ``rankone" comparator, therefore, it could be expected that the ``rankone" comparator and quality metric present the highest correlation of all participants.
    
    The ERC curves for the ``Wild vs Wild" scenario show that FaceQnet v0 struggles in the presence of bad quality images, where its performance is worse than all other algorithms participating in the evaluation. Again, this confirms the observations extracted based on the distributions shown in the top graph. Based on these results, we may say that the metric is able to detect poor images (see the high lobe close to 0 in the ``Wild" distribution), but it assigns to them always the same very low quality. Therefore, it needs to improve its ability to better discriminate between pictures corresponding to low values (quality range $0$-$30$).
\end{itemize}

The limitations handling low quality images of FaceQnet v0 revealed in this evaluation, have been partially addressed in the new release of the tool, FaceQnet v1, described in the present work, through: 1) a change in the architecture adding a dropout layer to avoid the quick saturation of the algorithm in the low-end of the quality range; and 2) an improvement of the training process using additional datasets and face recognition systems to produce the groundtruth quality scores. To evaluate the improvement in performance due to the changes introduced, FaceQnet v1 has been evaluated following a very similar protocol and metrics as those used in the NIST evaluation. This self-assessment is described in the following section. 

\subsection{Self-conducted Assessment: FaceQnet v1}
\label{selfevaluation}

In this evaluation we followed a testing protocol similar to the one of NIST described in the previous section. The target is to evaluate the improvements of the FaceQnet v1 model we have developed in the current work.

We tested the FaceQnet v1 model on $4$ different datasets: VGGFace2 (no overlap with the training set), BioSecure, CyberExtruder, and LFW. These databases were captured under different conditions and therefore they present different levels of variability. Examples of images of these databases with different levels of quality are shown in Fig.~\ref{VGGFace2}. A short description of the databases is included here (the description of VGGFace 2 is included in Sect.~\ref{proposed} as it was also used in the development stage):


\begin{figure}[t!]
\begin{center}
\includegraphics[width=\linewidth]{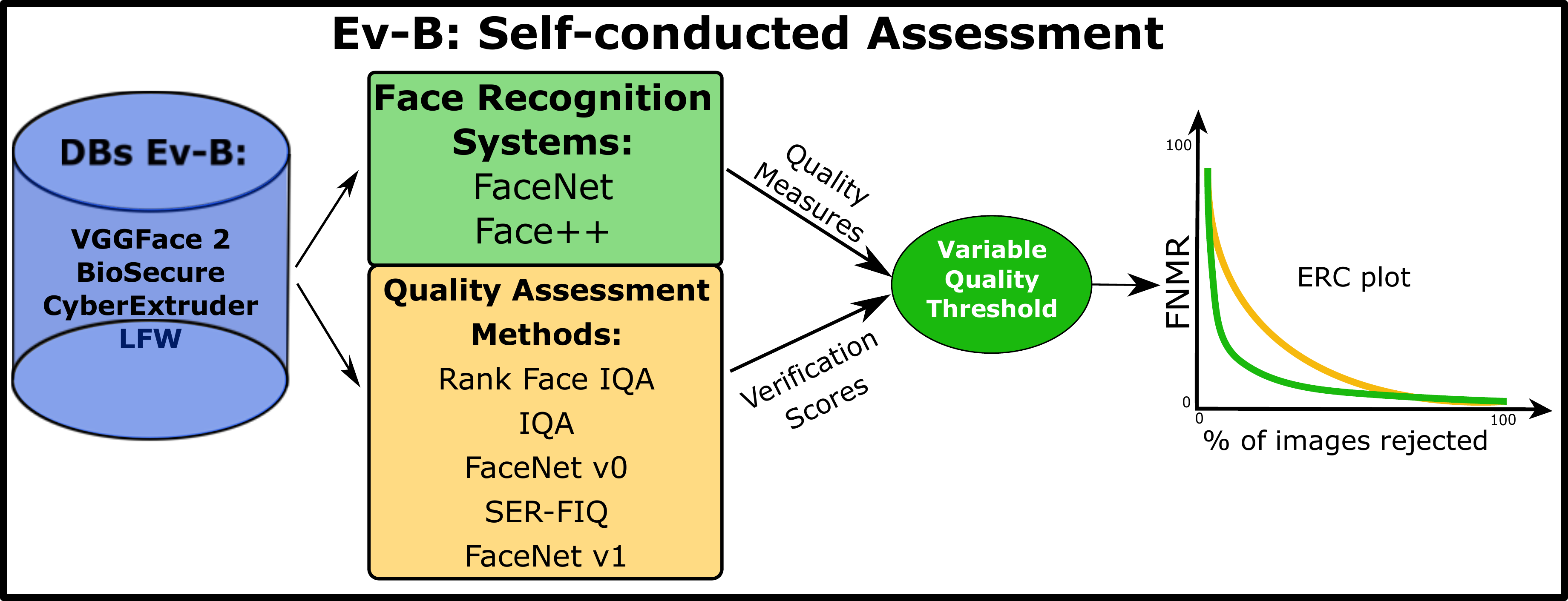}
\end{center}
\caption{\textbf{Experimental scheme for testing FaceQnet v1}. We computed only the mated verification scores for all the images in the test databases. In parallel, the quality of all the images involved in these mated pairs is obtained using FaceQnet v1. Finally, we calculated the FNMR values when discarding those mated pairs in which at least one of the images has its quality measure under a variable threshold. The mated comparison scores were computed using two different face recognition systems (FaceNet and Face++). The four test databases are: VGGFace2, BioSecure, CyberExtruder, and LFW.}
\label{validation_scheme}
\end{figure}

\begin{itemize}
    \item \textit{BioSecure Database (DB Ev-2B)}: 
The BioSecure Multimodal Database \cite{ortega2010multiscenario} consists of \mbox{$1$,$000$} subjects whose biometric samples were acquired in three different scenarios. Images for the first scenario were obtained remotely using a webcam, the second is a portrait-type scenario using a high quality camera with homogeneous background, and the third scenario is uncontrolled, captured with mobile cameras both indoors and outdoors. In the present work we have used this database for evaluation purposes. We used \mbox{$1$,$459$} images of $140$ subjects from the second and third scenarios for obtaining their quality measures with FaceQnet v1. 

    \item \textit{CyberExtruder Dataset (DB Ev-3B)}:
We used all the images in the CyberExtruder database\footnote{The Ultimate Face data set was provided by CyberExtruder.com, Inc. 1401 Valley Road, Wayne, New Jersey, 07470, USA} to perform the final accuracy tests of FaceQnet v1. The dataset contains \mbox{$10$,$205$} images of \mbox{$1$,$000$} subjects extracted from Internet. The data is unrestricted, i.e., it contains large pose, lighting, expression, race, and age variability. It also contains images with occlusions.
    
    \item \textit{Labeled Faces in the Wild (DB Ev-4B)}:
The Labeled Faces in the Wild (LFW) \cite{LFWTech} database has been also processed by FaceQnet v1 in order to label it with quality measures for accuracy tests. The database consists of \mbox{$13$,$233$} images of \mbox{$5$,$749$} different subjects, having \mbox{$1$,$680$} of them two or more different images available. This database has been widely used in recent years for studying face recognition under unconstrained conditions. Publishing an accuracy-based quality measure for each image can help to boost the accuracy of state-of-the-art face recognition systems that use this dataset for their benchmarks.
    
\end{itemize}


\begin{figure}[t!]
\begin{center}
\includegraphics[width=\linewidth]{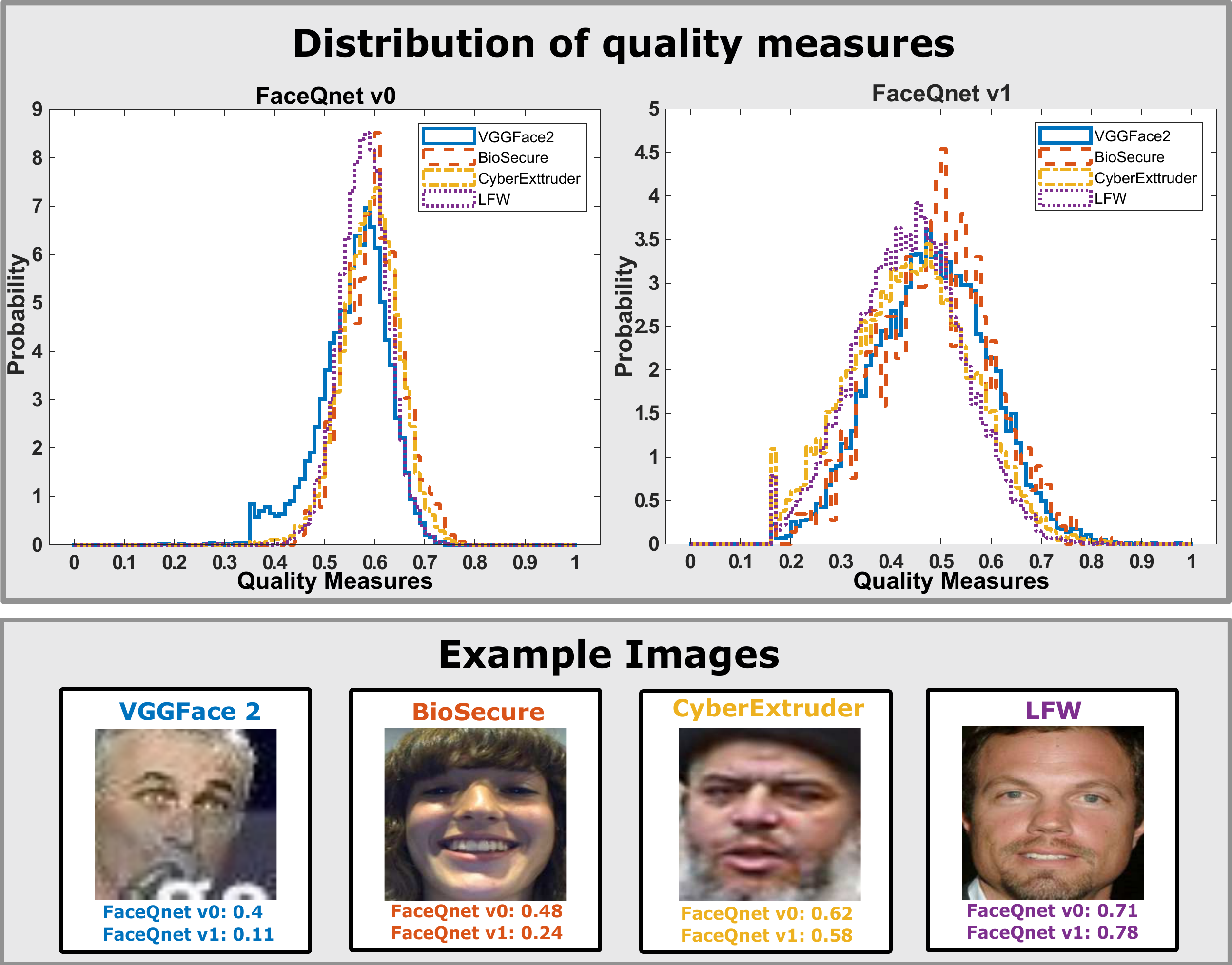}
\end{center}
\caption{\textbf{(Top) Distribution of the quality measures} for the VGGFace2, BioSecure, CyberExtruder, and LFW databases, obtained using FaceQnet v0 and FaceQnet v1. \textbf{(Bottom) sample images and their quality measures} for both models (v0 and v1). The example images illustrate how the new quality measures are more widespread along the [$0$,$1$] range than the old ones. VGGFace2 images obtained lower quality measures compared with those from the other databases since they contain more variability. The current FaceQnet v1 model distinguishes better between the quality of the different databases.}
\label{histograma}
\end{figure}

The experimental scheme for validating FaceQnet is shown in Fig.~\ref{validation_scheme}. First, we processed all the images from each test database with FaceQnet v1, obtaining a quality measure for each individual image. The resulting distributions of the quality measures are shown in Fig.~\ref{histograma}, for both FaceQnet v0 and v1. That figure also shows some example images and their associated quality measures. The results show that the scores obtained using FaceQnet v1 are more widespread along the [$0$,$1$] range than the ones obtained with FaceQnet v0. As expected, the VGGFace2 database presents a higher amount of low quality images since it represents real world acquisition conditions, while the quality values for the LFW or the BioSecure databases are slightly higher since their images were acquired in more controlled conditions.


\begin{figure}[t!]
\begin{center}
\includegraphics[width=\columnwidth]{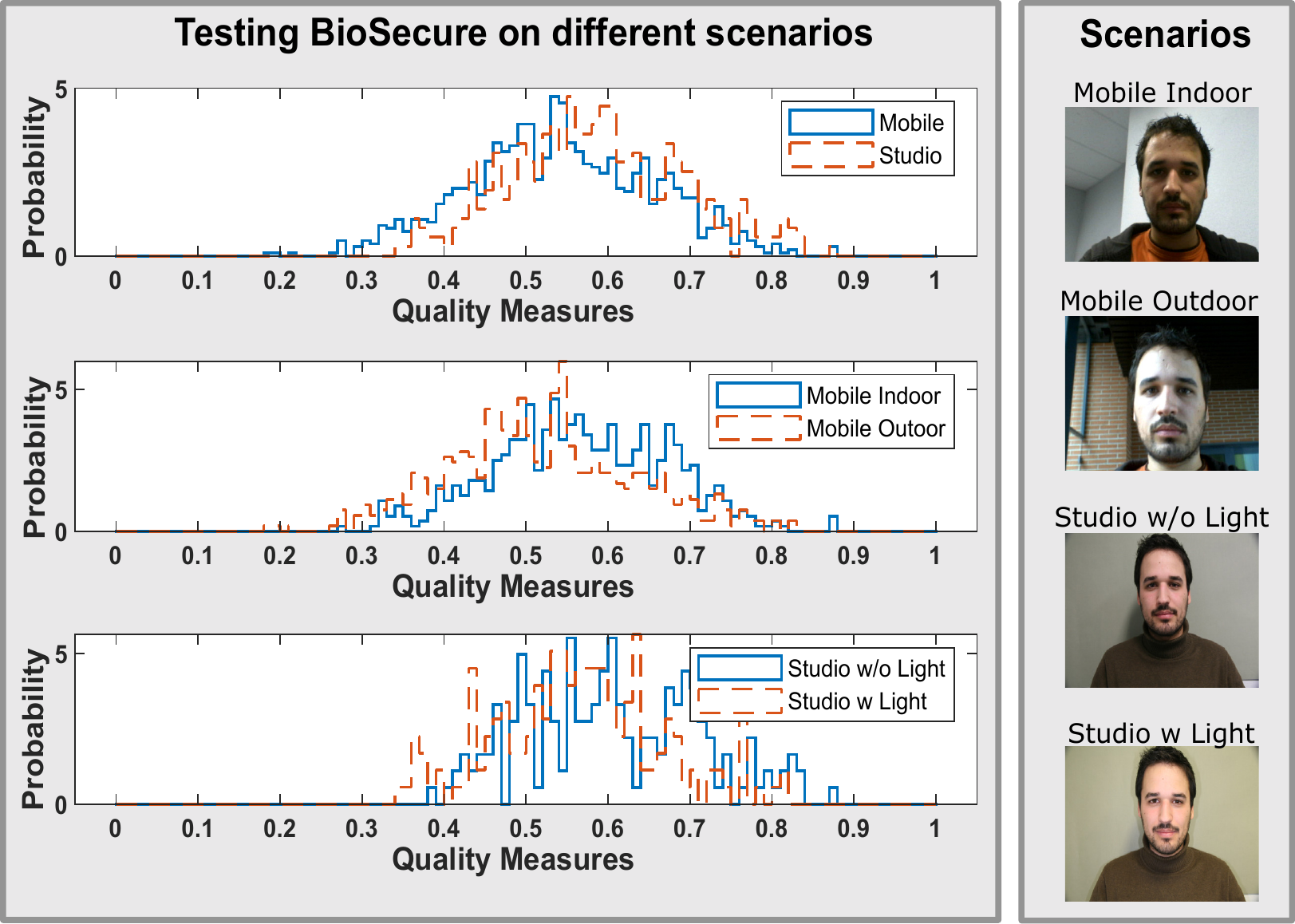}
\end{center}
\caption{\textbf{Distribution of the quality measures} for the different scenarios of the BioSecure database. The testing set of the database comprehends two different scenarios: mobile and studio. The mobile images are divided in indoor and outdoor subscenarios. The studio images were acquired with and without artificial illumination.}
\label{biosecure_scenarios}
\end{figure}

The testing dataset from BioSecure used in this self-evaluation has its images divided in two scenarios: 1) a portrait-type scenario acquired using a high quality camera, with homogeneous background, and under natural and artificial illumination conditions (referred to as ``studio'' scenario), and 2) an uncontrolled scenario, captured indoors and outdoors using mobile cameras (referred to as ``mobile'' scenario). This shapes a total of four subscenarios: ``studio with illumination'', ``studio without illumination'', ``mobile indoor'', and ``mobile outdoor''. 

We processed all the images of each one of the BioSecure subscenarios to determine if FaceQnet v1 is able to distinguish properly between the different types of images. Fig.~\ref{biosecure_scenarios} shows the distribution of the quality measures for the mentioned scenarios and subscenarios. As expected, the quality measures obtained for the ``studio'' conditions present a higher mean value than the ones from the ``mobile'' conditions, since its images were obtained with a camera of higher quality, with homogeneous background and illumination, better pose, etc. Additionally, the varying acquisition conditions of the ``mobile outdoor'' subscenario make the quality measures to be more spread along the quality range.

\subsection{Experimental Comparison to the State-of-the-Art}
\label{comparison_art}

In the last experiment of this self-evaluation we computed Error versus Reject Curves (similarly to the NIST evaluation) for comparing the accuracy of the quality measures obtained with FaceQnet v1 against other Quality Assessment (QA) methods. We decided to compare FaceQnet v1 with Rank Face IQA, a QA method for face recognition based on hand-crafted features \cite{chen2014face}. We also implemented a method designed for general Image Quality Assessment (IQA) \cite{idealods2018imagequalityassessment} to check how well it performs when applied to face images. Additionally, we evaluated SER-FIQ \cite{terhorst2020ser} as a representative of state-of-the art deep learning Face IQA methods. Finally, we included FaceQnet v0 \cite{2019_FaceQnet_Hernandez} in the comparison (the same version evaluated by NIST as described in the previous subsection).

\begin{figure}[t!]
\begin{center}
\includegraphics[width=\columnwidth]{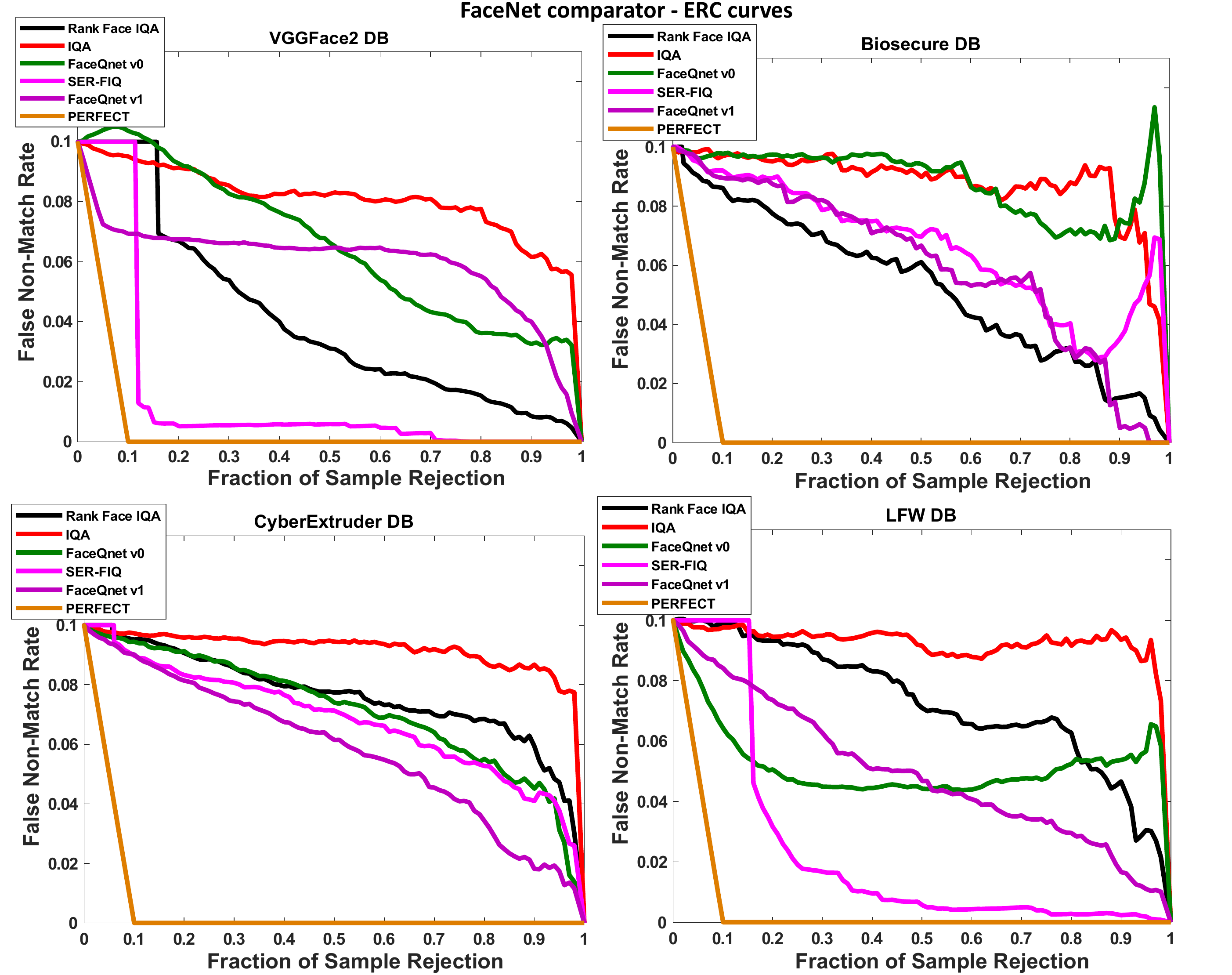}
\end{center}
\caption{\textbf{ERC obtained with the FaceNet comparator over the four evaluation datasets.} The initial FNMR has been set to 10\%. Fractions of the images with lowest quality measures have been removed consecutively. Five different QA algorithms have been used for obtaining quality measures of the testing images: a general Image Quality Assessment (IQA) method \cite{idealods2018imagequalityassessment}, a method for face QA based on hand-crafted features \cite{chen2014face}, SER-FIQ~\cite{terhorst2020face}, FaceQnet v0 \cite{2019_FaceQnet_Hernandez}, and the FaceQnet model of this paper (FaceQnet v1). The line labelled PERFECT is generated using $\max(\textrm{FNMR} - \textrm{Fraction of Sample Rejection} , 0)$. The closer the quality algorithm line is to the PERFECT line, the more related the quality measure is to face recognition accuracy. (Color image)}
\label{ERC_facenet}
\end{figure}

In this case we compute ERC plots for two different comparators, one of them used also during the development phase, i.e., FaceNet (SYSTEM Ev-1B FaceNet in Fig.~\ref{big_figure}), and a COTS recogniser never seen before by FaceQnet v1: Face++ from MEGVII \cite{megviface} (SYSTEM Ev-2B Face++ in Fig.~\ref{big_figure}). We decided to use Face++ because we wanted to check the accuracy of the quality measures from FaceQnet v1 against a commercial face recogniser not seen during the development stage. Additionally, it is easy and direct to use, since it consists of a cloud-based free API, making unnecessary having a powerful computer for obtaining the verification scores. Face++ performs a comparison between two face images returning a numerical comparison score between $0$ and $100$, while FaceNet returns a value between $0$ and $1$. In both cases, the higher the score, the higher the probability of a mated comparison. 


\begin{figure}[t!]
\begin{center}
\includegraphics[width=\columnwidth]{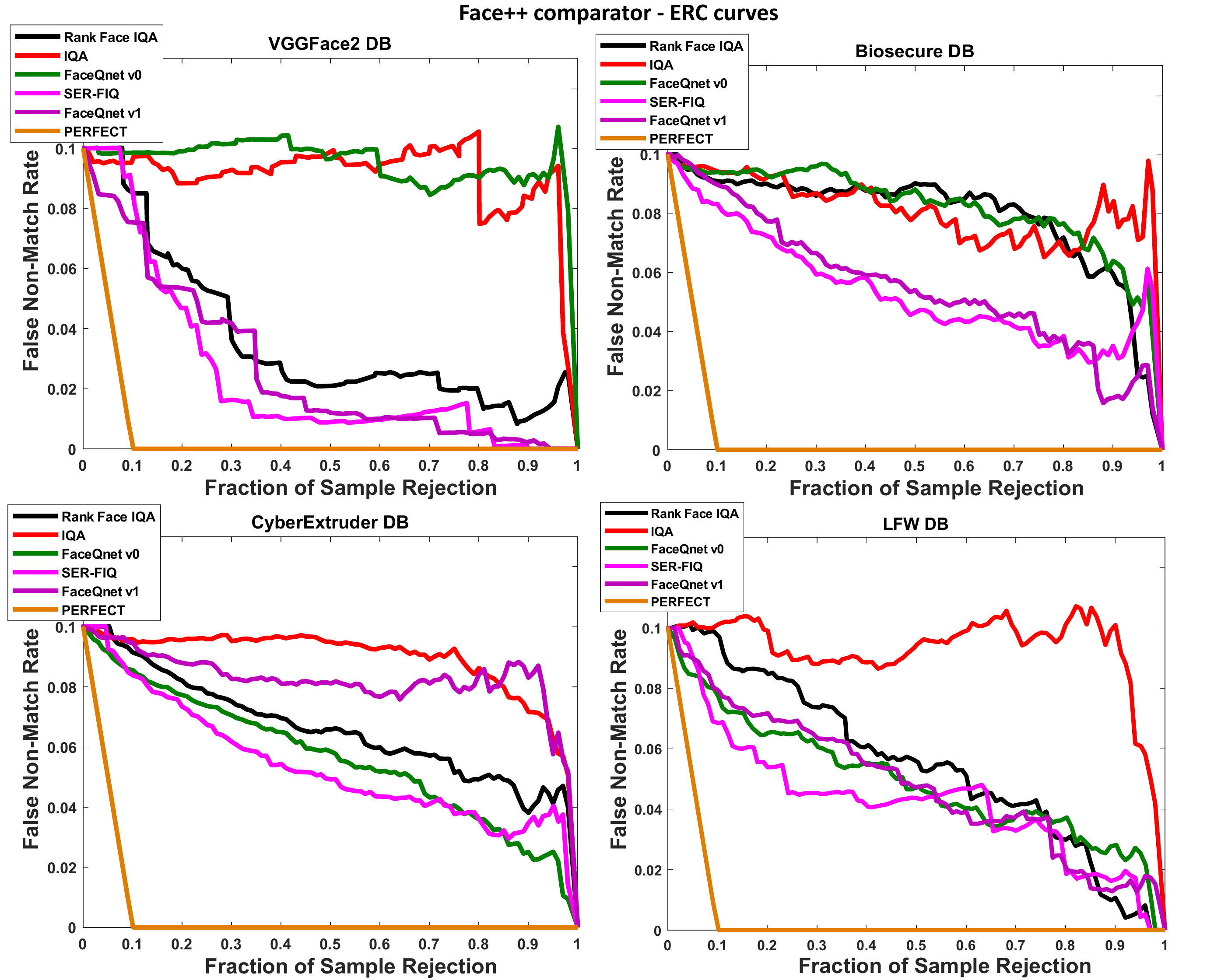}
\end{center}
\caption{\textbf{ERC obtained with the Face++ comparator over the four evaluation datasets.} The experimental protocol is the same used for FaceNet (see Figure \ref{ERC_facenet}). The line labelled PERFECT is generated using $\max(\textrm{FNMR} - \textrm{Fraction of Sample Rejection} , 0)$. The closer the quality algorithm line is to the PERFECT line, the more related the quality measure is to face recognition accuracy. (Color image)}
\label{ERC_faceplusplus}
\end{figure}

We compute the ERC plots for each combination of one testing database and one face recogniser from the evaluation set. In Figs.~\ref{ERC_facenet} and \ref{ERC_faceplusplus} we have fixed the verification thresholds to obtain an initial FNMR of 10\% for all the recognisers when using all the images indistinctly. 


Regarding the results, the FaceQnet model trained in this paper (FaceQnet v1) is always among the quality assessment methods with higher correlation with the face recognition performance. SER-FIQ and Rank Face IQA also obtain good results in their quality predictions. 

For FaceNet (see Fig.~\ref{ERC_facenet}), SER-FIQ obtains the best global results for VGGFace2 and LFW databases when discarding images of low quality. However, its behaviour is not consistent along all the quality range, not obtaining improvement when discarding the fraction of images with worst quality (i.e. the lowest $10$\%). Compared to SER-FIQ, FaceQnet v1 does not achieve so good global results in those two databases, but it obtains a highest improvement of performance when discarding that $10$\% of lowest quality images. 

For Face++ (see Fig.~\ref{ERC_faceplusplus}), FaceQnet v1 also stays among the best QA methods, except in the case of the CyberExtruder database where FaceQnet v0 and SER-FIQ present a higher correlation between their quality measures and the recognition performance. 

The general IQA algorithm from \cite{idealods2018imagequalityassessment} slightly increases the accuracy of the face recognisers when discarding low quality images. However, its performance is quite poor when compared with the other QA methods that have been adjusted specifically for face quality assessment. This algorithm has been designed for detecting variability factors such as blur, resolution, and homogeneity, but looking at the complete image. These are factors that can affect the accuracy of face recognition, but they might not be the most relevant to detect which images are actually suitable for face recognition. The face QA methods are focused on the zone of the image that contains the face to be analysed. The method from \cite{chen2014face} obtained good results in face QA but, due to its hand-crafted nature, it might perform worse when facing data from other databases and/or scenarios. It would be difficult to adjust this algorithm against different types of images and variability factors. On the other hand, FaceQnet has the potential to be easily adjustable to any possible scenario using a set of training images for fine-tuning the deep model.

After analysing all the ERC plots, it can be stated that FaceQnet v1 generates quality measures generally more correlated with the accuracy of face recognition compared to FaceQnet v0, showing also a consistent behaviour along the complete quality range.

\section{Conclusion and Discussion}
\label{conclusion_section}

The unattainable dream of so many human endeavours: knowing now, what is awaiting tomorrow. But, is it really an unreachable goal? In different contexts, mathematical models are getting more and more accurate at this seemingly impossible task. This is the case of biometric quality metrics. In a way, biometric quality is a window into the future. Even if this can seem a too-poetic of a definition for a computer algorithm, in reality, it may not be that farfetched. Biometric quality metrics allow to have an estimation in the present, of the accuracy that a system will achieve in the future on some given set of data. It is not difficult to grasp the huge value of a tool capable of such a feat.

The importance of assessing data quality for improving the performance of operational systems has been long known among the biometric community. Already in 2006 and 2007 the US NIST organised two back-to-back workshops\footnote{https://www.nist.gov/itl/iad/image-group/biometric-quality-workshop-i, https://www.nist.gov/itl/iad/image-group/biometric-quality-workshop-ii}, exclusively dedicated to the discussion of biometric quality and the promotion of research in the field. 

As a result of these and other similar initiatives, quality estimation algorithms are being increasingly deployed worldwide. Large national and international IT systems such as the US-VISIT, US Personal Identity Verification (PIV), or the EU Visa Information System (VIS) and Schengen Information System (SIS), mandate the measurement and reporting of quality scores of captured images. This is already being achieved on a regular basis in the case of fingerprints, where there has been a huge effort dedicated to the study of quality metrics. This investment has paid off great dividends, and has led to the development of NFIQ 2.0, a system-independent, open-source fingerprint quality metric which has been included as the common quality benchmark in the ISO/IEC 29794 standard. 

In spite of its importance, unanimously agreed by biometricians, the field of biometric quality assessment is far less advanced in most biometric characteristics than in the case of fingerprints. This is the situation where face recognition finds itself at the moment.

As recently as 2018, the US NIST organised a dedicated workshop to discuss all aspects related to face recognition technology\footnote{https://www.nist.gov/news-events/events/2018/11/international-face-performance-conference-ifpc-2018}, open to all interested parties, including academia, governmental institutions, law-enforcement agencies, border management agencies, and industry. Among the conclusions of the event, one of the urgencies identified by all stakeholders was to address the lack of reliable face quality metrics by fostering research in this underdeveloped field. An analogue conclusion was reached in 2019 by the European Commission, following their study for the integration of an automated face recognition system in the Schengen Information System and in other large European IT systems \cite{galbally2019study}. The report, aimed at policy makers, echoes the appeal made by multiple law enforcement entities in Europe for the development of a standard, system-independent, face quality metric similar to the existing NFIQ2 in fingerprint recognition.

This rapidly spreading awareness of the blatant lack of sufficient investment in face quality, has triggered a number of international initiatives to address the problem. Among them, the FRVT Quality Assessment campaign held by US NIST\footnote{https://pages.nist.gov/frvt/html/frvt\_quality.html}, is the first evaluation campaign aimed at comparing face quality metrics and set the current state of the art in the field, that will allow us to understand the strengths and limitations of existing technology. Another example of the international commitment to tackle this issue, is the launch by ISO/IEC JTC 1 SC37, the committee for standardisation in biometrics, of a collaborative work item in face quality, with the ambitious objective of producing standard algorithms for face quality estimation\footnote{https://www.iso.org/standard/78488.html?browse=tc}. 

While the commitment from international institutions and policy makers is an essential part of the equation, real advance in face quality metrics requires fuel. Ultimately, research is the driving force at the core of all this technology. With this pressing necessity as main motivation, the present work can be regarded as a solid contribution to bridge the existing gap in face quality, advancing the field by producing: an up-to-date overall picture of the state of the art, new insights, new open source algorithms, reproducible results following standard evaluation protocols, and new public data for future research and advancement.

In particular, we have developed FaceQnet v1, a new quality metric powered by deep learning technology which receives as input a face image and produces a scalar quality score as an estimation of the suitability of the picture to be used within face recognition systems. As a mean to the collective effort being made to advance the domain of face quality, FaceQnet v1 is put at the disposal of the community as an open source tool through GitHub\footnote{https://github.com/uam-biometrics/FaceQnet}, together with the quality scores produced for each of the four test datasets used in the evaluation (VGGFace2, BioSecure, CyberExtruder, and LFW).

In order to reach the most meaningful conclusions possible, FaceQnet has undergone a double evaluation: 

\begin{itemize}
    
    \item \textit{US NIST independent assessment}. The initial version of FaceQnet (v0) was submitted with the first wave of algorithms to the on-going FRVT-QA evaluation campaign organised by US NIST. In that evaluation, while showing promising results with respect to the other participants, the original algorithm revealed some of its flaws. In the current work these limitations have been partially corrected with a new version of FaceQnet (v1).
    
    \item \textit{Self-assessment}. We have carried out a reproducible self-assessment of the metric, based on public data and following a standard evaluation protocol. This evaluation has shown an improvement with respect to the preliminary version of the algorithm presented in \cite{2019_FaceQnet_Hernandez} (the same submitted to the NIST evaluation). The new metric described in this work has corrected some of the existing weak points, such as overfitting and system dependence, following a modification in the architecture and the training process. This evaluation has also shown the competitiveness of FaceQnet with respect to other state-of-the-art algorithms.
    
\end{itemize}

There are two major and, in our opinion, very valuable conclusions that can be drawn from these evaluations, regarding the two hypotheses that have been made in the work:

\begin{itemize}

\item \textbf{CONCLUSION 1}. Hypothesis 1 is confirmed. The approach followed in the present work for generating groundtruth quality scores, holds. It is safe to assume that the comparison score between a perfect quality picture A (i.e., ICAO compliant picture) and a picture B of lower quality (of the same subject), is a valid and accurate reflection of the quality level of picture B. Therefore, the comparison score thus produced, can be used as a machine-generated groundtruth quality score for picture B. This strategy allows automatising the groundtruth generation process, avoiding the highly time- and resource-consuming task of producing quality scores based on human perception, which may also be biased with respect to machines’ understanding of quality~\cite{2020_AAAI_Discrimination_Serna}.
    
\item \textbf{CONCLUSION 2}. Hypothesis 2 is confirmed. Machine-learned features for face recognition contain, not only the information regarding the \textit{identity} of the person, but also the information regarding the \textit{quality} of the picture. This quality-related information can be extracted from the original feature vector through a knowledge transfer process. Therefore, we can conclude that quality and identity are not only linked at the score level (quality measures are predictors of mated scores), but also at feature level. This new piece of knowledge, we believe can be very impactful and of high added value for the face quality forum, as the amount of labelled data available for face recognition is far higher than that tagged for face quality analysis. Subsequently, it is possible to accurately train from scratch Deep Neural Networks (DNNs) for face recognition (or use one of the already trained models), while, on the other hand, such a process may not be feasible at the moment for face quality estimation. However, the confirmation of hypothesis 2 allows us to overcome this scarcity of data, releasing the full potential of deep learning systems developed for face recognition to be applied as well in quality estimation tasks.

\end{itemize}

In addition to the two lessons learned pointed out above, the experimental evaluation of FaceQnet has also disclosed some critical points in the design of the algorithm, that need to be carefully taken into account if a similar approach is applied by other researchers for the development of face quality metrics, most importantly:

\begin{itemize}
\item \textit{Selection of the training ICAO-compliant images}. One of the key points in the present approach is the generation of the groundtruth quality scores based on a perfect ICAO compliant picture (see hypothesis 1). Due to the lack of public databases specifically designed for face quality assessment, such ICAO portraits were selected from an all-purpose face database, relying on an automated ICAO-compliance tester which efficiency has not been sufficiently proven. A manual supervision of the automatically selected pictures was performed as a second check, in order to ensure, to the largest extent possible, an overall high-quality level. In spite of our best efforts, it is likely that many of those training images, even though of high quality, were not fully ICAO compliant. Therefore, it is our strong believe that the training process would largely benefit if it was carried out on images initially acquired under ICAO restrictions, and not selected from a general ``in-the-wild" type database.\\

\item \textit{Training database}. A popular machine learning principle preaches: ``in God we trust, all others must bring data". Or, in other words, the more data, the better. Only more accurate results can be expected if the size of the training database is significantly increased, with images that cover substantially and uniformly the whole quality spectrum. This means that the training database should comprise for \textit{each single subject}: 1) pictures acquired in an ICAO compliant environment (see bullet point above); 2) pictures covering a large range of quality values (e.g., close-to-ICAO, frontal webcam indoor, frontal webcam outdoor, in the wild). To the best of our knowledge, there is still not such a database available to the research community. This would be, in our view, an invaluable asset in order to further advance the field of face quality assessment. 
    
\item \textit{Face detector}. In order to avoid biased results derived from features extracted from the background (i.e., if the background is homogeneous, the image may be automatically classified as ICAO), \textit{face} quality assessment algorithms should rely exclusively on information stemming from the \textit{face}, separating in this way the task of face detection from the task of face quality assessment. This means that, for an input image, the first task is to detect \textit{only} the face and to perform a tight crop solely of that area in the picture. This way the face detector may have difficulties in properly locating the face, but that difficulty would be independent from the face-only quality metrics that we advocate for. This is the most flexible and informative approach for dealing with biometric quality in general, but we understand that in some applications using a quality metric that integrates both the face segmentation and the biometric-only quality may be more efficient and operational. Although in our vision the biometric segmentation (face detection in this case) is not intrinsically part of the face quality algorithm, it can have a decisive impact on its outcome, depending on the accuracy of the face detector utilised. For the training and evaluation of face quality metrics, including FaceQnet, it is highly recommended to use face images with groundtruth segmentation for the face area, so that a face detector is not required and, therefore, the possible variability introduced by it is removed from the system.
    
\end{itemize}

Even though our work has been developed particularly in the framework of face biometrics, the proposed methodology for building a fully automated quality metric can be useful for other problems as well. Our methods can in fact be the basis to develop performance prediction tools for any automated artificial intelligence pipeline when dealing with a specific input.

As a wrap-up, we can say that the present work represents a step forward in the arduous quest for the generation of robust, system-independent, standard face quality metrics. All algorithms, results, and data described in the article have been made available to the community, so that this work can serve as a cornerstone to further advance this fundamental field, for the future deployment and development of face recognition technology.

After all, let’s not forget that, as we stated at the beginning of this article, the results of a computerised system are only as reliable as the data you input. If you input data that is garbage, the result will be unreliable garbage. Consequently, detecting garbage at the input, should be a compulsory critical task for any automated system.

\section*{Acknowledgment}
This work was supported in part by projects BIBECA from MINECO/FEDER (RTI2018-101248-B-I00), IDEA-FAST (IMI2-2018-15-two-stage-853981), PRIMA (ITN-2019-860315), and TRESPASS-ETN (ITN-2019-860813). The work was conducted in part during a research stay of J. H.-O. at the Joint Research Centre, Ispra, Italy. He is also supported by a PhD Scholarship from UAM. We also thank the experimental support given by Rudolf Haraksim at the JRC.


\bibliography{mybibfile}

\end{document}